\documentclass[sigconf]{acmart}

\usepackage{booktabs} 
\usepackage{amsmath}
\usepackage{caption}
\usepackage{subcaption}
\usepackage{enumitem}
\usepackage{comment}
\usepackage{subcaption}
\usepackage{graphicx}
\usepackage{bm}
\usepackage{mathrsfs}
\usepackage{amsthm}
\usepackage{threeparttable}
\usepackage{multirow}
\usepackage{xurl} 
\usepackage{balance}
\newtheorem{theorem}{Theorem}[section]

\newtheorem{definition}[theorem]{Definition}

\newtheorem*{conjecture*}{Conjecture}
\newtheoremstyle{nonindented}{1ex}{1ex}{}{}{\bfseries}{.}{.5em}{}
\newtheoremstyle{indented}{1ex}{1ex}{\itshape\addtolength{\leftskip}{0.6cm}\addtolength{\rightskip}{0.6cm}}{}{\bfseries}{.}{.5em}{}
\theoremstyle{nonindented}
\theoremstyle{indented}
\theoremstyle{plain}

\renewcommand{\hat}{\widehat}
\renewcommand{\tilde}{\widetilde}
\renewcommand{\bar}{\overline}

\newcommand{\eat}[1]{}

\newenvironment{lp*}{\begin{equation*}  \begin{array}{lll}}{\end{array}\end{equation*}}

\definecolor{applegreen}{rgb}{0.55, 0.71, 0.0}

\copyrightyear{2022}
\acmYear{2022}
\setcopyright{acmcopyright}
\acmConference[WSDM '22]{Proceedings of the Fifteenth ACM International Conference on Web Search and Data Mining}{February 21--25, 2022}{Tempe, AZ, USA}
\acmBooktitle{Proceedings of the Fifteenth ACM International Conference on Web Search and Data Mining (WSDM '22), February 21--25, 2022, Tempe, AZ, USA}
\acmPrice{15.00}
\acmDOI{10.1145/3488560.3498391}
\acmISBN{978-1-4503-9132-0/22/02}

\begin{document}
\fancyhead{}

\title{Learning Fair Node Representations with \\ Graph Counterfactual Fairness}

\author{Jing Ma$^1$, Ruocheng Guo$^2$,  Mengting Wan$^3$, Longqi Yang$^3$, Aidong Zhang$^1$, Jundong Li$^1$}\authornote{\label{note1}Corresponding Author.}

\affiliation{
  \institution{$^1$University of Virginia, Charlottesville, VA \country{USA} 22904\\ $^2$City University of Hong Kong, Hong Kong SAR \country{China}\\
  $^3$Microsoft, Redmond, WA\country{USA} 98052}
     \{jm3mr, aidong, jundong\}@virginia.edu, ruocheng.guo@cityu.edu.hk, \{mengting.wan, Longqi.Yang\}@microsoft.com
}

\renewcommand{\shortauthors}{}

\newcommand{\mymodel}{GEAR}
\newcommand{\bigCI}{\mathrel{\text{\scalebox{1.07}{$\perp\mkern-10mu\perp$}}}}
\begin{abstract}
Fair machine learning aims to mitigate the biases of model predictions against certain subpopulations regarding sensitive attributes such as race and gender. Among the many existing fairness notions, counterfactual fairness measures the model fairness from a causal perspective by comparing the predictions of each individual from the original data and the counterfactuals. {In counterfactuals, the sensitive attribute values of this individual had been modified.} Recently, a few works extend counterfactual fairness to graph data, but most of them neglect the following {facts that can lead to biases: 1) the sensitive attributes of each node’s neighbors may causally affect the prediction w.r.t. this node; 2) the sensitive attributes may causally affect other features and the graph structure. }
To tackle these issues, in this paper, we propose a novel fairness notion -- graph counterfactual fairness, which considers the biases led by the above facts. To learn node representations towards graph counterfactual fairness, we propose a novel 
framework based on counterfactual data augmentation. In this framework, we generate counterfactuals corresponding to perturbations on each node's and their neighbors' sensitive attributes. 
Then we enforce fairness by minimizing the discrepancy between the representations learned from the original graph and the counterfactuals for each node. 
Experiments on both synthetic and real-world graphs show that our framework outperforms the state-of-the-art baselines in graph counterfactual fairness, and also achieves comparable prediction performance. 
\end{abstract}

%



\maketitle

{\fontsize{8pt}{8pt} \selectfont
\textbf{ACM Reference Format:}\\
Jing Ma, Ruocheng Guo, Mengting Wan, Longqi Yang, Aidong Zhang,
Jundong Li. 2022. Learning Fair Node Representations with Graph Counter-
factual Fairness. In \textit{Proceedings of the Fifteenth ACM International Conference on Web Search and Data Mining (WSDM ’22), February 21–25, 2022, Tempe, AZ, USA.} ACM, New York, NY, USA, 9 pages. https://doi.org/10.1145/3488560.3498\\391}

\section{Introduction}
\begin{figure}[t]
\centering
\begin{subfigure}[b]{0.235\textwidth}
        \centering
        \includegraphics[height=1.1in]{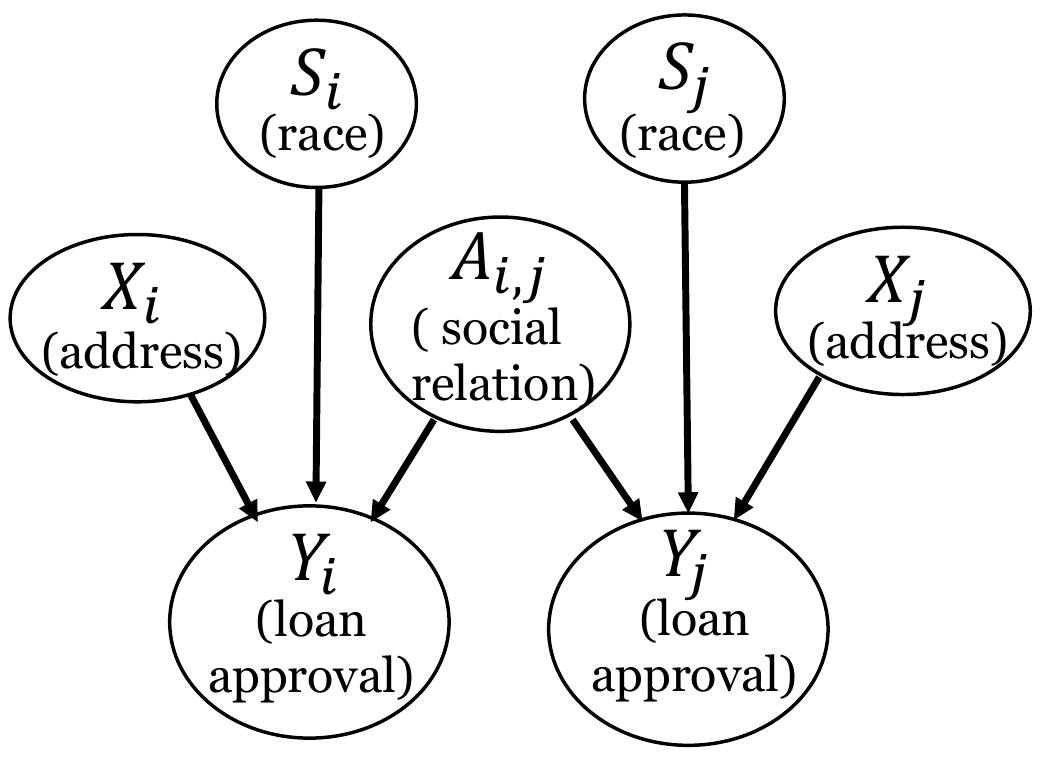}
        \caption{$M'$}
    \end{subfigure}
  \begin{subfigure}[b]{0.235\textwidth}
        \centering
        \includegraphics[height=1.1in]{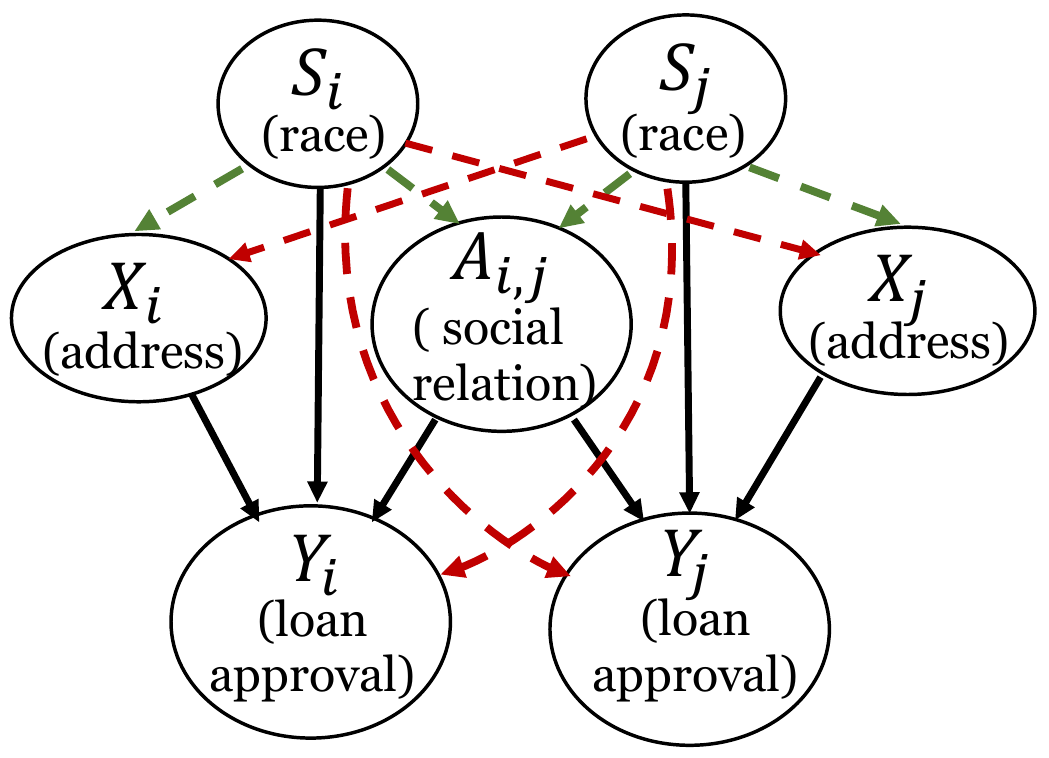}
        \caption{$M$}
    \end{subfigure}
    \vspace{-6mm}
    \caption{Causal models generally used in existing works ($M'$) and in this work ($M$). We use $S_i,X_i, Y_i$ to denote the sensitive attribute, features, and label of any node $i$, and $A_{i,j}\in\{0,1\}$ denotes the edge between node pair $(i,j)$. Each arrow denotes a causal relation. The dashed lines denote the causal relations that the existing works do not consider.}
    \vspace{-5mm}
     \label{fig:causal}
 \end{figure}
 
Representation learning on graphs aims to map nodes into a latent embedding space. These node representations are often used to power downstream predictive tasks, and have become the new state-of-the-art in multiple real-world applications \cite{wu2020comprehensive,kipf2016semi,velivckovic2017graph,xu2018powerful}. However, these node representation learning approaches may overlook potential biases buried in the graph data, thus introducing algorithmic biases against subpopulations defined by certain sensitive attributes such as race, gender, and age. Consequently this may raise ethical and societal concerns, especially in high-stake decision-making scenarios such as {ranking of job applicants \cite{mehrabi2021survey} and credit scoring \cite{sekhon2016perceptions}}. For example, it would become a serious ethical issue if a bank's decision on the loan application was affected by the applicant's and their close contacts' race information.
 

To tackle the above problem, several approaches were proposed to assess and address the fairness of node representation learning on graphs. The majority of these methods aim to learn node representations which can elicit \textit{statistically} fair predictions across the population \cite{bose2019compositional,agarwal2021towards,dai2020fairgnn,li2020dyadic}. In addition, the concept of \textit{counterfactual fairness} has been extended to graph-structured data recently \cite{agarwal2021towards,agarwal2021towards2}. Different from the previous statistical notions, counterfactual fairness extends Pearl's causal structural models \cite{pearl2009causal} and aims to encourage the predictions made from different versions of the same individual (a.k.a. \textit{counterfactuals}) to be equal. For example, the prediction for one's loan application being approved should be the same regardless this applicant being Black or White.


This paper falls under the umbrella of counterfactual fairness but focuses on addressing two critical limitations of existing studies \cite{agarwal2021towards,agarwal2021towards2} of counterfactual fairness on graphs: 1) biases induced by \textbf{one's neighboring nodes} and 2) biases induced by \textbf{the causal relations from the sensitive attributes to other features as well as the graph structure}. We follow the previous loan application example to explain these limitations in details: i) As illustrated in Fig.~\ref{fig:causal}(a), existing studies mostly focus on mitigating the causal influence from the sensitive attribute (race information $S_i$) of the $i$-th applicant on the prediction of the label (loan approval decision $Y_i$), but neglect the fact that the race information of the applicant's social contacts ($S_j$) can also causally affect the fairness of the prediction (as labeled using {\color{red} red} dashed edges in Fig.~\ref{fig:causal}(b)). 
ii) On the other hand, existing methods may implicitly assume the sensitive attribute ($S_i$) has no causal effect on other variables such as node features ($X_i$) and the graph structure ($A_{i,j}$) so that they can safely simplify the counterfactual data generation mechanism as by just flipping the sensitive attribute values. 
However, we question the applicability of this assumption since such causal effect is ubiquitous in real-world scenarios. For example, one's race 
can causally influence their social relations as well as the residential neighborhood they live in (as labeled using {\color{applegreen} green} dashed edges in Fig.~\ref{fig:causal}(b)).\footnote{There may also exist causal relations between non-sensitive features and the graph structure, although we do not show them in Fig.\ref{fig:causal} for simplicity of illustration.}

We argue that biases in model predictions can be induced by the aforementioned pathways.  
In this paper, we propose a more comprehensive fairness notion on graphs -- \textit{graph counterfactual fairness}, 
which considers the potential biases regarding the sensitive attributes of each node and its neighboring nodes, as well as the biases led by the causal effect  from sensitive attributes on other variables. 
%
%
With this notion, learning node representations towards graph counterfactual fairness is still challenging. It is because the causal relations among  variables (as showed in Fig.~\ref{fig:causal}(b)) are often required to obtain the counterfactuals, but these causal relations are often unknown in practice. 
Manually constructing the entire causal model requires extensive domain knowledge and human efforts, 
especially for large-scale graph data.
To address the above challenge, we propose a novel framework to learn \textbf{G}raph count\textbf{E}rfactually f\textbf{A}ir node \textbf{R}epresentations (\textbf{\mymodel}). \mymodel~ aims to learn node  representations towards graph counterfactual fairness, and maintain high performance for downstream tasks such as node classification.
\mymodel~ includes the following modules: 
    1) \textbf{Subgraph generation.} 
    {To reduce the costs of modeling the causal relations on large graphs, }
    we first develop an algorithm to automatically infer the importance scores among nodes. For each individual node, we then prune the range of the causal model to an ego-centric subgraph which contains only the node itself and its most influential neighboring nodes. 
    2) \textbf{Counterfactual data augmentation.} For each node, we leverage the graph auto-encoder technique \cite{kipf2016variational} and fairness constraints to generate two types of counterfactuals as data augmentation: 1) self-perturbation: the counterfactuals where each node's own sensitive attribute value had been modified; 2) neighbor-perturbation: the counterfactuals where the sensitive attribute values of neighbors had been modified.
    3) \textbf{Node representation learning.} To learn node representations towards graph counterfactual fairness, 
    we leverage a Siamese network \cite{bromley1993signature} to minimize the discrepancy between the representations learned from the original subgraph and those learned from counterfactuals.
The main contributions of this work can be summarized as follows:
\begin{itemize}
\item \textbf{Problem.} We propose a new fairness notion --- graph counterfactual fairness, which considers the potential biases brought by different causal pathways from sensitive attributes to the graph model predictions.
%
\item \textbf{Method.} We propose a novel framework \mymodel~ to learn node representations towards graph counterfactual fairness. Specifically, for each node, we minimize the discrepancy between the representations learned from the original data and the augmented counterfactuals with different sensitive attribute values.
\item \textbf{Experiments.} We conduct extensive experiments on both synthetic and real-world graphs.
The results show that the proposed method outperforms existing baselines in multiple fairness notions, and achieves comparable prediction performance in downstream tasks.
\end{itemize}
\section{Problem Definition}
\begin{table}[t]
\small
  \caption{Notation.}
  \label{tab:notation}
  \vspace{-4mm}
  \begin{tabular}{ll}
    \toprule
    Notation & Definition \\ 
    \midrule
    $\mathcal{G}$,$\mathcal{V},\mathcal{E}$          & the original graph, the set of vertices/edges \\
    $\mathbf{X},\mathbf{x}_i$ & features of all nodes/the $i$-th node\\
    $\mathbf{A}$          & adjacency matrix\\
    $n$               & the number of nodes \\
    $\mathbf{S},{s}_i$ & sensitive attribute values of all nodes/the $i$-th node\\
    $\mathbf{Z},\mathbf{z}_i$ & representations of all nodes/the $i$-th node\\
    $\Phi(\cdot),\phi(\cdot)$  & encoder/subgraph encoder\\
    $\mathtt{AGG(\cdot)}$  & aggregator \\
    $f(\cdot)$             & downstream classifier/predictor \\
    $(U)_{V \leftarrow v}$ & counterfactual of variable $U$ when $V$ had been set to $v$\\
    $\neg i$          & the indices which are not $i$\\
    $d,d'$               & dimension of features/representations \\
    $(\cdot)^{(i)}$   & information of the subgraph with central node $i$ \\
    $\mathtt{SUB}(\cdot)$ & subgraph generation operator \\
    $\mathtt{SMP}(\cdot)$ & sampling operation of sensitive attribute\\
    $\tilde{\mathbf{S}}$ & the summary of neighboring sensitive attribute values\\
    $\mathcal{\bar{G}}^{(i)}$,$\mathcal{\underline{G}}^{(i)}$  & the set of counterfactual subgraphs of $\mathcal{G}^{(i)}$ under \\ & self-perturbation/neighbor-perturbation \\
    $C$           & the sampling number in neighbor-perturbation\\
    \bottomrule
\end{tabular}
\vspace{-2mm}
\end{table}
\textbf{Notations.} Given a graph $\mathcal{G}=\{\mathcal{V}, \mathcal{E}, \mathbf{X}\}$, where $\mathcal{V}$ is the set of nodes, $\mathcal{E}$ is the set of edges, $\mathbf{X}=\{\mathbf{x}_i\}_{i=1}^n$ denotes the node features $(n=|\mathcal{V}|)$, and $\mathbf{x}_i \in \mathbb{R}^d$ represents the features of node $i$. $\mathbf{A}\in \mathbb{R}^{n\times n}$ denotes the adjacency matrix of the graph $\mathcal{G}$, where $\mathbf{A}_{i,j}=1$ if edge $i\rightarrow j$ exists, otherwise $\mathbf{A}_{i,j}=0$. Without loss of generalization, we assume $\mathcal{G}$ is undirected and unweighted, but this work can be naturally extended to directed or weighted settings. 
Each node $i$ has a sensitive attribute ${s}_i \in \{0, 1\}$ (we assume one single, binary sensitive attribute for simplicity, but our model can also be easily extended to multivariate or continuous sensitive attributes). $\mathbf{S}=\{s_i\}_{i=1}^n$, and ${s}_i$ is included in $\mathbf{x}_i$. We denote the non-sensitive features as $\mathbf{X}^{\neg s}=\{\mathbf{x}^{\neg s}_1,...,\mathbf{x}^{\neg s}_n\}$, where $\mathbf{x}^{\neg s}_i = \mathbf{x}_i \backslash {s}_i$.

Traditional node representation learning methods train an encoder  $\Phi(\cdot): \mathbb{R}^{n\times d} \times \mathbb{R}^{n \times n}\rightarrow \mathbb{R}^{n\times d'}$ to map each node to a latent representation. The learned representations for the $n$ nodes are denoted by $\mathbf{Z}=\{\mathbf{z}_i\}_{i=1}^n$, where $\mathbf{z}_i=(\Phi(\mathbf{X},\mathbf{A}))_i$, $\mathbf{z}_i\in \mathbb{R}^{d'}$ for any node $i$, and $d'$ is the dimensionality of node representations. These representations can be used in various downstream tasks like node classification \cite{bhagat2011node}, link prediction \cite{liben2007link}, and graph classification \cite{ying2018hierarchical}. $f(\cdot)$ denotes the downstream classifier/predictor. In the node classification task, let $y_i$ denote the true label of the node $i$, $f(\cdot)$ takes the representation $\mathbf{z}_i$ as input, and outputs the predicted label $\hat{y}_i$.

%
\noindent\textbf{Counterfactual fairness.} Counterfactual fairness \cite{kusner2017counterfactual} is a fairness notion based on Pearl's structural causal model \cite{pearl2009causal}. A \textit{causal model} consists of  a \textit{causal graph} and \textit{structural equations}. A causal graph is a directed acyclic graph (DAG), where each node represents a variable, and each directed edge represents a causal relationship. Structural equations describe these causal relations among  variables. For variables $Y, S$, the value of the \textit{counterfactual} "what would $Y$ have been if $S$ had been set to $s$?" is denoted by $Y_{S\leftarrow s}$. 
Based on a given causal model, a predictor $\hat{Y}=f(X)$ is \textit{counterfactually fair} \cite{kusner2017counterfactual} if under any features $X=x$ and sensitive attribute $S=s$, 
\vspace{-1mm}
\begin{equation}
P(\hat{Y}_{S\leftarrow s}=y|X=x,S=s) = P(\hat{Y}_{S\leftarrow s'}=y|X=x,S=s),
\vspace{-1mm}
\end{equation}
for all $y$ and $s'\ne s$. Here $\hat{Y}_{S\leftarrow s}=f(X_{S\leftarrow s},s)$ denotes the prediction made on the counterfactual when $S$ had been set to $s$. Intuitively, it aims to minimize the difference between predictions made on each individual and its counterfactuals with different sensitive attribute values. Ideally, the counterfactuals should be generated based on the ground truth causal model. 
Different from the statistical fairness notions such as equality of opportunity (EO) \cite{hardt2016equality,zafar2017fairness} and demographic parity (DP) \cite{zemel2013learning}, counterfactual fairness 
aims to eliminate the biases led by the causal effect from the sensitive attribute on the observed variables used for model training. 
However, most existing works of counterfactual fairness focus on i.i.d. data.

\noindent\textbf{Existing notion of counterfactual fairness on graph.} 
Recent works \cite{agarwal2021towards,agarwal2021towards2} have extended counterfactual fairness to graphs.   
Given a graph $X=\mathbf{X}, A=\mathbf{A}$, these works consider that an encoder $\Phi(\cdot)$ satisfies  counterfactual fairness if for any node $i$:
\vspace{-1mm}
\begin{equation}
    (\Phi({X}_{{S}_i={0}},{A}))_i = (\Phi({X}_{{S}_i= {1}},{A}))_i,
    \label{eq:naive_gcf}
    \vspace{-1mm}
\end{equation}
where ${X}_{{S}_i= {0}}$ and ${X}_{{S}_i= {1}}$ denote the node features after setting $S_i$ as $0$ and $1$, respectively, while everything else does not change\footnote{We use italicized uppercase letters (e.g., $S_i,X,A$) to denote random variables, and use 
italicized lowercase letters (e.g., $s_i$), non-italicized bold lowercase/uppercase letters (e.g., $\mathbf{x}_i$ and $\mathbf{X}$) to denote specific realization of scalars or vectors/matrices, respectively. }.
This notion considers fairness as minimizing the discrepancy between the representations of each node with different values of its sensitive attribute (while everything else is fixed). This notion has the following limitations: 1) 
it does not consider the potential biases led by the causal effect from the sensitive attribute of other nodes in the graph on the prediction of each node; 2) it implicitly assumes that the sensitive attribute has no causal effect on other features or the graph structure. 
In a nutshell, this fairness notion is more limited than the general counterfactual fairness notion. 

\noindent\textbf{Graph counterfactual fairness.} To address the above limitations, in this work, we propose a novel fairness notion on graphs: 
\vspace{-2mm}
\begin{definition}
{(Graph counterfactual fairness).}
An encoder $\Phi(\cdot)$ satisfies graph counterfactual fairness if for any node $i$:
\begin{equation}
    P(({Z}_i)_{{S}\leftarrow \mathbf{s}'}|X=\mathbf{X},A=\mathbf{A})=P((Z_i)_{{S}\leftarrow \mathbf{s}''}|X=\mathbf{X},A=\mathbf{A}),
    \label{eq:gcf}
\end{equation}
\noindent for all $\mathbf{s}' \!\ne \!\mathbf{s}''$, where $\mathbf{s}', \mathbf{s}''\! \in\! \{0,1\}^{n}$ are arbitrary sensitive attribute values of all nodes, $Z_i=(\Phi(X,A))_i$ denotes the node representations for node $i$. In other words, given a graph $X=\mathbf{X}, A=\mathbf{A}$, $\Phi(\cdot)$  
should minimize the distribution discrepancy between the representations $(\Phi({X}_{{S}\leftarrow \mathbf{s}'},{A}_{{S}\leftarrow \mathbf{s}'}))_i$ and $(\Phi({X}_{{S}\leftarrow \mathbf{s}''},{A}_{{S}\leftarrow \mathbf{s}''}))_i$ for any node $i$. 
\end{definition}
Intuitively, this notion encourages the representations learned from the original graph and counterfactuals to be equal. The counterfactuals correspond to different cases when the sensitive attribute of the $n$ nodes had been set to any values. 
For notation simplicity, in the following sections, we use {$\mathbf{X}_{{S}\leftarrow \mathbf{s}'}$ to denote a specific value of the counterfactual ``what would the node features have been if the sensitive attribute of the $n$ nodes had been set by $\mathbf{s}'$, given the original data, i.e., node features $\mathbf{X}$ and graph structure $\mathbf{A}$?"}.  We also use notation  $\mathbf{A}_{{S}\leftarrow \mathbf{s}'}$ in a similar way.

In this work, we aim to develop a framework which learns node representations on graph towards graph counterfactual fairness, and maintains a good prediction performance simultaneously.

\section{The Proposed Framework --- GEAR}
\begin{figure*}[t]
\centering
     \includegraphics[width=.82\textwidth,height=2.8in]{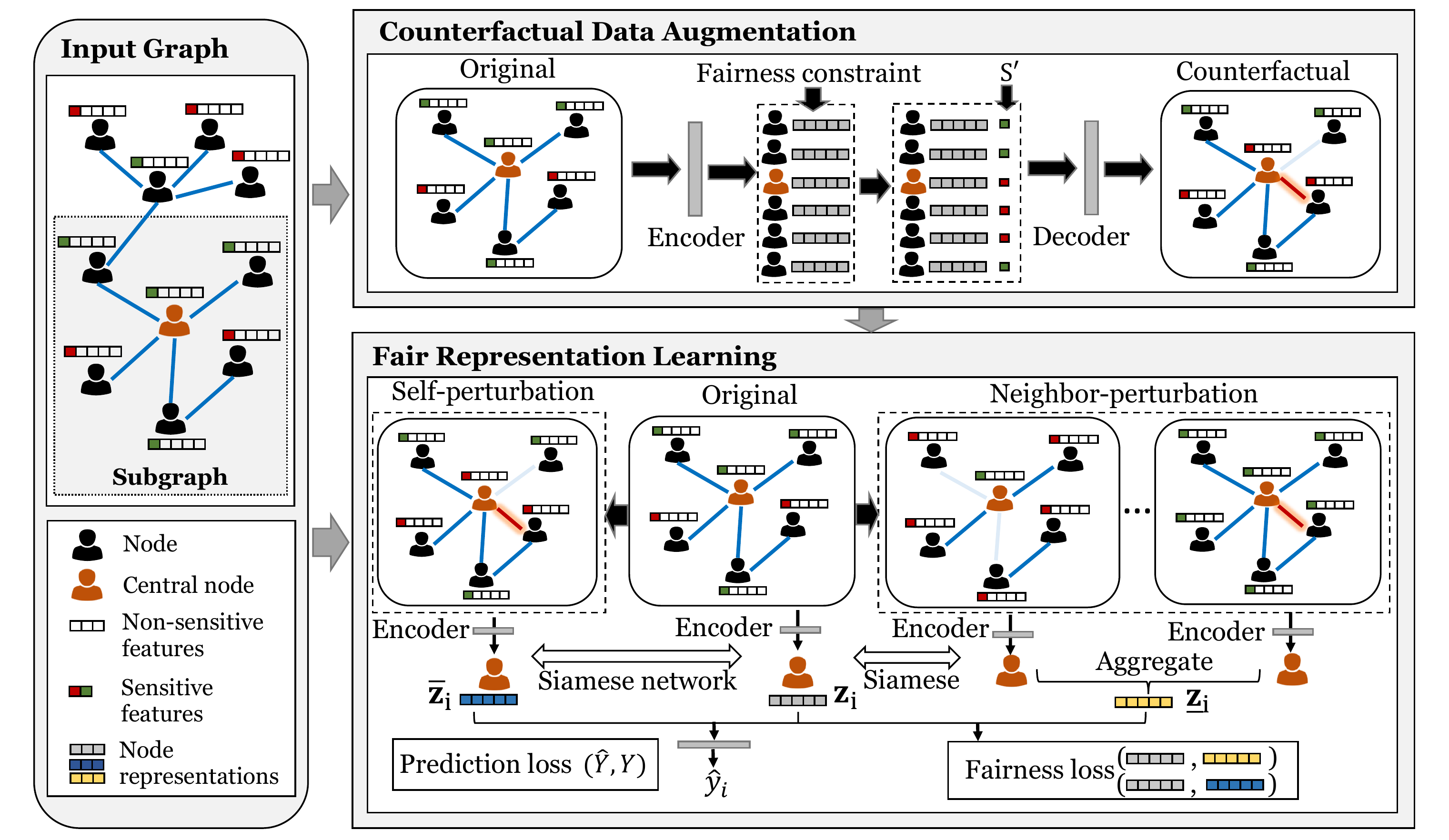}
     \vspace{-0.15in}
      \caption{An illustration of the proposed framework \mymodel.}
      \vspace{-0.1in}
       \label{fig:model_1}
\end{figure*}

In this section, we propose a novel framework \mymodel~ which aims to learn node representations for graph counterfactual fairness. As the illustration shown in Fig.~\ref{fig:model_1}, \mymodel~ mainly includes three key components: 1) subgraph generation; 2) counterfactual data augmentation; 3) fair representation learning. In subgraph generation, \mymodel~ extracts a context subgraph for each node, which contains the local graph structure including the node itself (central node) and its nearest neighbors with respect to precomputed importance scores.
In counterfactual data augmentation, we generate counterfactuals in which the sensitive attribute of nodes in these subgraphs had been perturbed. 
%
Based on the augmented counterfactuals, the fair representation learning component leverages Siamese networks \cite{bromley1993signature} to minimize the distance between the representations learned from the original data and the counterfactuals w.r.t. the same node. 


\vspace{-2mm}
\subsection{Subgraph Generation}
True causal models for graph data are often difficult to be completely obtained, 
especially for large-scale graphs.
Based on a common observation \cite{hamilton2017inductive,jiao2020sub} that each node is mostly influenced by its nearest neighbors, we extract a subgraph $\mathcal{G}^{(i)}$ with node features $\mathbf{X}^{(i)}$ and adjacency matrix $\mathbf{A}^{(i)}$ for each node $i$.
This subgraph extracts the context information of the central node $i$ on $\mathcal{G}$, i.e., the subgraph of $\mathcal{G}$ which only contains the top $k$ neighbors of node $i$ (including itself). These top-$k$ neighbors are usually within several hops from the central node. 
%
Specifically, for each node $i$ on the graph, we generate its context subgraph $\mathcal{G}^{(i)}$ with a subgraph generator $\mathtt{Sub}(\cdot)$. 
Based on these context subgraphs, we learn the representations for their corresponding central nodes. This is based on a commonly used assumption \cite{hamilton2017inductive} that each node has a low dependency with the nodes outside its context subgraph. Therefore, each subgraph is expected to be informative enough with respect to the graph structure relative to the central node for high-quality representation learning and counterfactual data augmentation afterwards.
%

Inspired by recent subgraph based node representation learning methods \cite{jiao2020sub,zhang2020graph}, we first compute the importance scores for every node pair with personalized pagerank algorithm \cite{jeh2003scaling}. The importance scores can be calculated as:
$
\mathbf{R}= \alpha (\mathbf{I}-(1-\alpha)\bar{\mathbf{A}}),
$
where $\mathbf{R}$ is the importance score matrix, and each entry $\mathbf{R}_{i,j}$ describes how important node $j$ is for node $i$, and $\mathbf{R}_{i,:}$ denotes the importance score vector for node $i$.
$\alpha$ is a parameter in the range of $[0,1]$, $\mathbf{I}$ is the identity matrix. $\bar{\mathbf{A}}=\mathbf{A}\mathbf{D}^{-1}$ denotes the column-normalized adjacency matrix, where $\mathbf{D}$ is the corresponding diagonal matrix with $\mathbf{D}_{i,i}=\sum_j{\mathbf{A}_{i,j}}$.
We compute $\mathbf{R}$ in a preprocessing stage before model training for efficiency.
With the importance scores, we use a $\mathtt{TOP(\cdot)}$ operation to select the top-$k$ important nodes $\mathcal{V}^{(i)}$ for each central node $i$, then formulate the context subgraph $\mathcal{G}^{(i)}$ as follows:
\begin{equation}
    \mathcal{G}^{(i)} = \{\mathcal{V}^{(i)},\mathcal{E}^{(i)}, \mathbf{X}^{(i)}\} = \{\mathbf{A}^{(i)}, \mathbf{X}^{(i)}\},
    \label{eq:subgraph_gen}
\end{equation}
\begin{equation}
 \mathcal{V}^{(i)} = \mathtt{TOP}(\mathbf{R}_{i,:}, k),
\end{equation}
\begin{equation}
 \mathbf{A}^{(i)} = \mathbf{A}_{\mathcal{V}^{(i)}, \mathcal{V}^{(i)}}, \,\,\mathbf{X}^{(i)} = \mathbf{X}_{\mathcal{V}^{(i)},:},
 \label{eq:subgraph_gen_feature}
\end{equation}
where the symbol $:$ means all the indices. The above subgraph generation process (Eq.~\eqref{eq:subgraph_gen} to~\eqref{eq:subgraph_gen_feature}) is defined as $\mathcal{G}^{(i)}=\mathtt{Sub}(i, \mathcal{G}, k)$. Then the generated subgraphs are fed into encoders to learn representations of the central nodes.

\subsection{Counterfactual Data Augmentation} 
\label{subsec:cf_da}
To achieve graph counterfactual fairness, we pretrain a counterfactual data augmentation module before node representation learning.
Here we consider a relatively simple but general causal model (as shown in Fig.~\ref{fig:causal}(b)) to generate counterfactuals for each subgraph. 
Based on common observations \cite{kusner2017counterfactual}, we assume that the sensitive attribute (e.g., race) is exogenous, i.e., it has no parent variables in the causal graph, and it would causally influence the other node features, the graph structure, and the labels. 
Based on the causal model we assume, once we intervene on the sensitive attribute, we need to model how the other variables change accordingly. To achieve this goal, we use a graph variational auto-encoder (GraphVAE) \cite{kipf2016variational} based module, which takes each context subgraph as input and encodes each node in the subgraph into a latent embedding $\mathbf{h}_i$, then a decoder reconstructs the original subgraph with the latent embeddings $\mathbf{H}=\{\mathbf{h}_1,...,\mathbf{h}_k\}$ and the sensitive attribute values of the $k$ nodes in this subgraph. The reconstruction loss $\mathcal{L}_r$ is as follows (we leave out the superscript $(\cdot)^{(i)}$ for notation simplicity):
\begin{equation}
    \mathcal{L}_r = \mathbb{E}_{q({H}|\mathbf{X},\mathbf{A})}[-\log(p(\mathbf{X},\mathbf{A}|H, \mathbf{S}))] + \text{KL}[q({H}|\mathbf{X},\mathbf{A})\|p({H})],
\end{equation}
where 
$p({H})$ is a standard Normal prior distribution. We sample the embeddings $\mathbf{H}$ from $q({H}|\mathbf{X},\mathbf{A})$. 

As the sensitive attribute is assumed to be exogenous, we can mitigate the causal effect from the sensitive attribute on the embeddings by removing the statistical dependency between them. To achieve this target, we use an adversarial learning method to learn embeddings which are invariant to different sensitive attribute values of each node and their neighbors. Specifically, we use a discriminator here to predict the summary of neighboring sensitive attribute values. Here we take the summary $\tilde{s}_i$ as the mean aggregation over all the nodes in the subgraph $\mathcal{G}^{(i)}$, i.e., $\tilde{s}_i=\frac{1}{|\mathcal{V}^{(i)}|}\sum_{j\in \mathcal{V}^{(i)}}s_j$. We divide the summary into $B$ ranges to formulate it as a multivariate classification task for the discriminator $D(\cdot)$. We use a fairness constraint as follows:
$
    \mathcal{L}_d = \sum\nolimits_{b\in [B]}\mathbb{E}[\log(D(\mathbf{H}, b))],
$
where the discriminator $D(\mathbf{H}, b)$ predicts the probability of whether the summary of sensitive attribute values is in range $b$. Based on the theoretic analysis in \cite{song2020novel,bose2019compositional}, $\mathcal{L}_d$ is a regularizer to minimize the mutual information between the summary of sensitive attribute values and the embeddings. The final loss of the counterfactual data augmentation is:
$
    \mathcal{L}_a = \mathcal{L}_r + \beta \mathcal{L}_d,
$
where $\beta$ is a hyperparameter for the weight of fairness constraint. We use alternating stochastic gradient descent for optimization: 1) we minimize $\mathcal{L}_a$ by fixing the discriminator and updating parameters in other parts; 2) we minimize $-\mathcal{L}_a$ with respect to the discriminator while other parts fixed.
%
To achieve graph counterfactual fairness, we expect the embeddings  $\mathbf{H}$ can capture the latent variables 
which are informative of the input subgraph but not causally influenced by the sensitive attribute of the nodes in the subgraph.
We pretrain the counterfactual data augmentation module to better disentangle different components of the framework. If more prior knowledge of the causal model is provided, we can incorporate it in counterfactual data augmentation, e.g., directly generate counterfactuals with a given causal model, and do not need to change other components in the framework.

Based on the above techniques, we conduct perturbations on the original subgraphs and obtain different types of counterfactuals. For each context subgraph $\mathcal{G}^{(i)}$, we generate two kinds of perturbations on it, including self-perturbation on the sensitive attribute of the central node, and neighbor-perturbation on the sensitive attribute of other nodes in the subgraph.

\noindent\textbf{Self-perturbation.} In the subgraph $\mathcal{G}^{(i)}$, we take its node embeddings, flip the sensitive attribute value of the central node ${s}_i$, then feed the embeddings and the perturbed sensitive attribute into the decoder of the pretrained counterfactual data augmentation module, and take the reconstructed subgraph as the corresponding counterfactual.
The set containing the subgraphs after self-perturbation is denoted by $\mathcal{\bar{G}}^{(i)}=\{\mathcal{G}^{(i)}_{{S}_i\leftarrow {1-s_i}}$\}. 

\noindent\textbf{Neighbor-perturbation.} Similarly, in the subgraph $\mathcal{G}^{(i)}$, we randomly perturb the sensitive attribute values of any nodes except the central node, i.e., the nodes in the set $\mathcal{V}^{(i)}_{\neg i}$. After such perturbation, we generate a set of counterfactuals $\mathcal{\underline{G}}^{(i)}=\{\mathcal{G}^{(i)}_{{S}^{(i)}_{\neg i}\leftarrow{\mathtt{SMP}({\mathbf{S}^{(i)}_{\neg i}})}}\}$, where $\mathtt{SMP}(\cdot)$ randomly samples specific values of the sensitive attribute out of the value space $\{0,1\}^{|\mathcal{V}^{(i)}|-1}$. 
We use a parameter $C$ to denote the number of  $\mathtt{SMP}(\cdot)$ operations in neighbor-perturbation.


\vspace{-2mm}
\subsection{Fair Representation Learning}
Based on the above counterfactual data augmentation, we learn fair representations which are expected to elicit the same predicted label across different counterfactuals w.r.t. the same node. 
To achieve this goal, we leverage Siamese networks \cite{bromley1993signature} to encode the three kinds of subgraphs: original subgraphs $\mathcal{G}^{(i)}$, counterfactual subgraphs $\mathcal{\bar{G}}^{(i)}$ and $\mathcal{\underline{G}}^{(i)}$ for each central node $i$. For graph counterfactual fairness, we expect to learn the same representations for each central node from the three kinds of subgraphs. 
We train a subgraph encoder $\phi(\cdot)$ to generate the representations $\mathbf{z}_i$, $\mathbf{\bar{z}}_i$, $\mathbf{\underline{z}}_i$ for each central node $i$ on these three kinds of subgraphs, respectively. Then we  minimize the distance between the central node representations learned from the original subgraph and from the counterfactuals. We formulate the loss for graph counterfactual fairness as:
\begin{equation}
    \mathcal{L}_f = \frac{1}{|\mathcal{V}|}\sum\nolimits_{i\in \mathcal{V}}((1-\lambda_s) d(\mathbf{z}_i, \mathbf{\bar{z}}_i) + \lambda_s d(\mathbf{z}_i, \mathbf{\underline{z}}_i)),
\end{equation}
where $d(\cdot)$ is a distance metric such as cosine distance. $\lambda_s\in [0,1]$ is a hyperparameter which controls the weight of neighbor-perturbation. From the original subgraph and the counterfactuals, we obtain the node representations in the following way: 
\begin{equation}
    \mathbf{z}_i 
    =(\phi(\mathbf{X}^{(i)}, \mathbf{A}^{(i)}))_i,
\end{equation}
\begin{equation}
    \mathbf{\bar{z}}_i  
    =\mathtt{AGG}(\{(\phi(\mathbf{X}_{{S}_i\leftarrow{1-s_i}}^{(i)}, \mathbf{A}_{S_i\leftarrow{1-s_i}}^{(i)}))_i\}),
\end{equation}
\begin{equation}
    \mathbf{\underline{z}}_i = 
    \mathtt{AGG}(\{(\phi(\mathbf{X}_{{S}^{(i)}_{\neg i}\leftarrow\mathtt{SMP}(\mathbf{S}^{(i)}_{\neg i})}^{(i)}, \mathbf{A}^{(i)}_{{S}^{(i)}_{\neg i}\leftarrow\mathtt{SMP}(\mathbf{S}^{(i)}_{\neg i})}))_i\}),
\end{equation}
where $\phi(\cdot): \mathbb{R}^{k\times d}\times \mathbb{R}^{k\times k} \rightarrow \mathbb{R}^{k\times d'}$ takes each subgraph as input, and embeds each node on the input subgraph into a latent representation.
We take the representations of each central node $i$ learned from the original data as $\mathbf{z}_i$, and we use $\mathbf{Z}=\{\mathbf{z}_i\}_{i=1}^n$ for downstream tasks. 
For the sampled counterfactual subgraphs in $\mathcal{\bar{G}}^{(i)}$ and $\mathcal{\underline{G}}^{(i)}$, we use an aggregator (e.g., mean aggregator) $\mathtt{AGG}(\cdot)$ to aggregate the representations of each central node $i$, and obtain the final representations $\mathbf{\bar{z}}_i$ and $\mathbf{\underline{z}}_i$. 

To encode useful information of node features and graph structure into the representations, we use labels as supervision. We use the task of node classification as an example, but our framework can be naturally extended to other kinds of tasks on graph data such as link prediction. We denote the class labels as $\mathbf{Y}=\{y_1,...,y_n\}$ for the $n$ nodes. The prediction loss can be formulated as:
\begin{equation}
    \mathcal{L}_p = \frac{1}{n}\sum\nolimits_{i\in[n]}l(f(\mathbf{z}_i), y_i),
\end{equation}
where $l(\cdot)$ is the loss function (e.g., cross-entropy) which measures the prediction error, $f(\cdot)$ makes predictions for downstream tasks with the representations, i.e., $\hat{y}_i=f(\mathbf{z}_i)$. Finally, the overall loss function for fair representation learning is:
\begin{equation}
    \mathcal{L} = \mathcal{L}_p + \lambda\mathcal{L}_f + \mu \|\theta\|^2,
\end{equation}
where $\theta$ is the set of model parameters, $\lambda$ and $\mu$ are hyperparameters controlling the weight of the graph counterfactual fairness constraint, and $L_2$ norm regularization, respectively.
\section{Experiments}
We evaluate the proposed method on both synthetic and real-world graphs. The detailed statistics of these datasets are shown in Table~\ref{tab:dataset}, including the number of nodes, the number of edges, the dimension of features, the average degree, and the number of intra-group and inter-group edges with respect to the sensitive attribute.
\begin{table}[t]
  \caption{Detailed statistics of the datasets.}
  \label{tab:datasets}
 \vspace{-3mm}
  \label{tab:dataset}
  \begin{tabular}{llll}
    \toprule
    Dataset & Synthetic & Bail & Credit\\
    \midrule
    $|\mathcal{V}|$ & $2,000$ & $18,876$ & $30,000$\\ 
    $|\mathcal{E}|$ & $4,120$ & $311,870$ & $137,377$\\ 
    Feature dimension & $26$ & $18$& $13$\\
    Average degree & $5.120$& $34.044$ &$10.158$\\
    \# of intra-group edges & $2,379$ & $162,821$& $120,750$\\
    \# of inter-group edges & $1,741$ &$149,049$& $16,627$\\
  \bottomrule
\end{tabular}
\vspace{-4mm}
\end{table}

\begin{table*}[t]
\small
\centering
 \caption{Comparison of the performance of node representation learning methods with respect to prediction and fairness.}
 \vspace{-2mm}
 \label{tab:baseline}
 \begin{tabular}{l||l||c|c|c||c|c|c|c}
 \hline
\multirow{2}{*}{Dataset} & \multirow{2}{*}{Method} & \multicolumn{3}{c||}{Prediction Performance} & \multicolumn{4}{c}{Fairness}                     \\\cline{3-9}
                         &                         & Accuracy ($\uparrow$)   & F1-score ($\uparrow$)   & AUROC ($\uparrow$)  & $\triangle_{EO}$ ($\downarrow$) & $\triangle_{DP}$ ($\downarrow$) & $\delta_{CF}$ ($\downarrow$) & $R^2$ ($\downarrow$)\\
\hline
Synthetic                   & GCN  &   $0.686 \pm 0.015$         &  $0.687 \pm 0.020$          &  $0.758 \pm 0.017$    &  $0.050 \pm 0.030$          &   $0.060 \pm 0.033$               &     \underline{$0.101 \pm 0.030$}          & $0.085 \pm 0.050$  \\
                         & GraphSAGE  &   \underline{$0.712 \pm 0.012$}         &  \underline{$0.714 \pm 0.021$}          &  \underline{$0.789 \pm 0.018$}    &   $0.049 \pm 0.036$         &   $0.053 \pm 0.042$               &   $0.172 \pm 0.056$ &   $0.011 \pm 0.011$           \\
                          & GIN   & $0.682 \pm 0.021$            &   $ 0.691 \pm 0.022$         &  $0.741 \pm 0.021$    &  $0.077 \pm 0.053$          &  $0.081 \pm 0.055$               &  $0.301 \pm 0.080$  &    $0.011 \pm 0.009$          \\
                         & C-ENC      &  $0.665 \pm 0.023$          & $0.671\pm 0.031$           & $0.732\pm 0.028$     &  \underline{$0.030\pm 0.024$}          &   \underline{$0.048\pm 0.026$}               &  $0.633\pm 0.013$     &$0.085\pm 0.016$           \\
                         & FairGNN  &  $0.668\pm 0.020$           &  $0.672\pm 0.026$          & $0.735\pm 0.022$     &  \bm{$0.025\pm 0.021$}          &   \bm{$0.042\pm 0.033$}               &   $0.678\pm 0.014$   &  $0.091\pm0.021$          \\
                         & NIFTY-GCN & $0.618 \pm 0.035$            &  $0.640 \pm 0.037$          & $0.672 \pm 0.042$     &   $ 0.172 \pm 0.110$         &   $0.199 \pm 0.106$               &   $0.208 \pm 0.090$     &  $0.105 \pm 0.081$        \\
                         & NIFTY-SAGE & $0.664 \pm 0.041$           &   $0.682 \pm 0.073$         &  $0.755 \pm 0.021$    &   $0.031 \pm 0.027$         &     $0.048 \pm 0.027$             &  $0.147 \pm 0.071$   &     \underline{$0.008 \pm 0.005$}        \\
                         & GEAR             &  \bm{$0.718 \pm 0.018$}          &    \bm{$0.724 \pm 0.022$}        &   \bm{$0.793 \pm 0.014$}    &    $0.052 \pm 0.038$        &  $0.064 \pm 0.038$                & \bm{$0.002 \pm 0.002$}          &  \bm{$0.007 \pm 0.006$}     \\
\hline
Bail               & GCN                     &   $0.838\pm 0.017$         &       $0.782\pm0.023$     &   $0.885 \pm 0.018$   &     $0.023 \pm 0.019$       &      $0.075 \pm 0.014$            &  $0.132 \pm 0.059$ &  $0.075 \pm 0.028$              \\
                         & GraphSAGE   &   \bm{$0.854 \pm 0.026$}         &   \bm{$0.804 \pm 0.032$}         &  \bm{$0.905 \pm 0.021$}    &  $0.039 \pm 0.022$          &  $0.086 \pm 0.039$                &  $0.088 \pm 0.047$       &   $0.069 \pm 0.011$      \\
                          & GIN                     &   $0.731 \pm 0.058$         &     $0.656 \pm 0.084$       &   $0.773 \pm 0.069$   &    $0.041 \pm 0.023$        &     $0.065 \pm 0.034$             & $0.143 \pm 0.069$   &    $0.047 \pm 0.036$          \\
                         & C-ENC  & $0.842 \pm 0.047$            &  $0.792\pm 0.014$          &  $0.889\pm 0.033$    & $0.038\pm 0.022$           &   $0.069\pm 0.020$               &    $0.040 \pm 0.025$      &  $0.078\pm 0.024$      \\
                         & FairGNN               &   $0.835\pm 0.024$    &    $0.784\pm 0.021$        & $0.882\pm 0.035$     &   $0.046\pm 0.013$         &   $0.074\pm 0.026$               &  $0.042 \pm 0.032$          &   $0.086 \pm 0.016$   \\
                          & NIFTY-GCN                   &  $0.752 \pm 0.065$          &  $0.669 \pm 0.050$          & $0.799 \pm 0.051$     &   \underline{$0.019 \pm 0.015$}         &     \bm{$0.036 \pm 0.022$}             &   $0.031 \pm 0.017$   &   $0.025 \pm 0.018$         \\
                         & NIFTY-SAGE                   &   $0.823 \pm 0.048$         &  $0.723 \pm 0.103$          &  $0.876 \pm 0.043$    &    \bm{$0.014 \pm 0.006$}        &  \underline{$0.047 \pm 0.015$}                & \underline{$0.013 \pm 0.011$}  &    \underline{$0.044 \pm 0.020$}           \\
         & GEAR             &     \underline{$0.852 \pm 0.026$}     &    \underline{$0.800 \pm 0.031$}        &  \underline{$0.896 \pm 0.016$}    &   \underline{$0.019 \pm 0.023$}         &     $0.058 \pm 0.017$             & \bm{$0.003 \pm 0.002$}    &   \bm{$0.038 \pm 0.012$}          \\
\hline
Credit                         & GCN                     &   $0.698 \pm 0.028$         &   $0.794 \pm 0.027$         &   $0.684 \pm 0.019$   &    $0.087 \pm 0.035$        &    $0.108 \pm 0.031$              &  $0.042 \pm 0.029$ &   $0.022 \pm 0.014$            \\
                         & GraphSAGE                     &  $0.739 \pm 0.009$          &  $0.821 \pm 0.008$          &  \bm{$0.756 \pm 0.011$}    &      $0.094 \pm 0.033$      &     $0.109 \pm 0.030$             &  $0.062 \pm 0.036$   &         $0.014 \pm 0.004$    \\
                          & GIN                     &    $ 0.713 \pm 0.018$        &   $0.805 \pm 0.016$         & $0.706 \pm 0.010$     &    $0.121 \pm 0.042$        &    $0.130 \pm 0.037$              &    $0.123 \pm 0.060$ &   $0.025 \pm 0.012$           \\
                         & C-ENC   &  $0.695\pm 0.011$          &    $0.786\pm 0.012$        & $0.683\pm 0.018$     &   $0.098\pm 0.025$         &  $0.104\pm 0.042$                & $0.100\pm 0.024$    &  $0.048\pm 0.012$           \\
                         & FairGNN                 &   $0.683\pm 0.053$         &   $0.780\pm 0.042$         &  $0.680\pm 0.021$    &    $0.175 \pm 0.035$        &   $0.187\pm 0.036$               &  $0.105\pm 0.053 $   &     $0.056 \pm 0.018$        \\
                         & NIFTY-GCN                   &   $0.697 \pm 0.007$         &  $0.792 \pm 0.007$          &   $0.685 \pm 0.007$   &   $0.097 \pm 0.024$         &  $0.106 \pm 0.021$                & $0.004 \pm 0.004$      &   $0.017 \pm 0.003$        \\
                         & NIFTY-SAGE                   &  \underline{$0.751 \pm 0.023$}          &      \underline{$0.833 \pm 0.020$}  &  $0.730 \pm 0.011$    &  \bm{$0.075 \pm 0.021$}          &   \bm{$0.094 \pm 0.019$}               &  \underline{$0.004 \pm 0.003$}     &     \underline{$0.011 \pm 0.003$}      \\
                         & GEAR            &   \bm{$0.755 \pm 0.011$}         &   \bm{$0.835 \pm 0.008$}         &   \underline{$0.740 \pm 0.008$}   &   \underline{$0.086 \pm 0.018$}         & \underline{$0.104 \pm 0.013$}       &   \bm{$0.001 \pm 0.001$}       &  \bm{$0.010 \pm 0.003$}\\
\hline
\end{tabular}
\vspace{-3mm}
\end{table*}

\vspace{-2mm}
\subsection{Datasets} 
\label{sec:dataset}
A synthetic dataset and two real-world datasets are used in the experiments. In the synthetic dataset, we create a causal model with which we can fully manipulate the data generation process. More specifically, in this synthetic dataset, we generate the features, latent embeddings, graph structure, and labels as below:
\begin{equation}
S_i\sim \text{Bernoulli}(p),\,\, {Z}_i\sim \mathcal{N}(0,\mathbf{I}),\,\, {X}_i=\mathcal{S}({Z}_i) + S_i\mathbf{v},\,\, 
\end{equation}
\begin{equation}
    P({A}_{i,j}=1)\!=\!\sigma(\text{cos}({Z}_i,{Z}_j)+ a\mathbf{1}({S}_i={S}_j)), Y_i=\mathcal{B}(\mathbf{w} Z_i + {w}_s \frac{\sum_{j\in \mathcal{N}_i}S_j}{|\mathcal{N}_i|}),
\end{equation}
where we sample the sensitive attribute with Bernoulli distribution, where $p=0.4$ is the probability of $S_i=1$. We sample latent embeddings ${Z}_i \in \mathbb{R}^{d_z}$ from a Gaussian distribution, where $d_z=50$. ${Z}_i$ influences the node features and the graph structure for each node $i$, and $\mathcal{S}(\cdot)$ denotes a sampling operation which randomly selects $d=25$ dimensions out of the latent embeddings to form the observed features ${X}_i$. 
$\mathbf{v}\in\mathbb{R}^d, \mathbf{v} \sim \mathcal{N}(0,\mathbf{I})$ controls the influence of the sensitive attribute on other features.
We simulate the probability of each edge $(i,j)$ based on the cosine similarity between ${Z}_i$ and ${Z}_j$, as well as whether their sensitive attribute values are equal. Here $\mathbf{1}(\cdot)$ is an indicator function which outputs $1$ when the input statement is true and $0$ otherwise. We set parameter $a=0.01$. 
Then we sum up the above similarity between $({Z}_i, {Z}_j)$ and the indicator function's output of $(S_i=S_j)$, and map it into a range of $[0,1]$ with a Sigmoid function $\sigma{(\cdot)}$ to compute the link probability between $(i,j)$. 
$\mathbf{w}\in \mathbb{R}^{d_z}$ contains parameters sampled from Normal distribution. We average each node's and their one-hop neighbors' sensitive attribute values and use it into label generation with weight ${w}_s=0.5$. In  $\mathcal{B}(\cdot)$, we map $Y_i$ into a binary value. Specifically, we first compute the mean value of $Y_i$ over all nodes, and set $Y_i=1$ if it is larger than the mean value, otherwise $Y_i=0$.

As for the real-world graphs, we use: 
1) \textit{Bail} \cite{agarwal2021towards}: This graph contains the data of defendants who got released on bail at the U.S state courts. 
In this graph, each node represents a defendant, each edge between a pair of nodes represents their similarity of criminal records and demographics. We use the defendants' race as the sensitive attribute. The task is to classify defendants into bail (not tend to commit a violent crime if released) or no bail. 
2) \textit{Credit defaulter} \cite{agarwal2021towards}: This graph contains people's default payment information. In this graph, each node represents an individual, each edge between a pair of nodes represents the similarity of their spending and payment patterns. We use their age as the sensitive attribute, and the task is to predict that their default ways of payment is credit card or not.

To evaluate the graph counterfactual fairness of the proposed method, we need to generate the ground-truth counterfactuals with the perturbations on different nodes' sensitive attribute. On the synthetic dataset, the counterfactuals can be generated based on the predefined causal model. On the real-world graphs, the ground-truth causal models are unknown, so we use a simple causal model and fit the observed data, and use the learned parameters in the fitted causal model to generate the counterfactuals for the whole graph. More specifically, we first use a Na\"ive Bayes model to learn $P(X_i|S_i)$, and then update the counterfactual features by $(\mathbf{x}_i)_{S_i\leftarrow{1-s_i}}=\mathbb{E}[{X}_i|S_i=1-s_i]-\mathbb{E}[{X}_i|S_i=s_i]+\mathbf{x}_i$. We use $(\cdot)^{CF}$ to denote any counterfactuals. 
Then we generate the counterfactual graph edge for each $(i, j)$ based on the following rules:
\begin{equation}
    P({A}^{CF}_{i,j}=1)= \sigma(\text{cos}({X}_i^{CF}\backslash S_i^{CF},{X}_j^{CF}\backslash S_j^{CF}) +  \gamma \mathbf{1}({S}^{CF}_i, {S}^{CF}_j)),
\end{equation}
where $\text{cos}(\cdot)$ is cosine similarity. We use $\sigma(\cdot)$ to map its input into a range $[0, 1]$ to compute the link probability for any node pairs in the counterfactual graph.
We fit the data with this causal model and learn the parameter $\gamma$. For evaluation, we use the counterfactual data generated by the causal model and learned parameters. 

As discussed in \cite{russell2017worlds}, there might be multiple possible causal models in the real-world data, so we have tried  different causal models to fit the real-world data. Due to the space limit, we only show the results based on the causal model as described above, but the observations over all the experiments are generally consistent.

\begin{table*}[t]
\small
\centering
 \caption{Comparison of the performance of different variants of \mymodel.}
 \vspace{-2mm}
 \label{tab:ablation}
 \begin{tabular}{l||l||c|c|c||c|c|c|c}
 \hline
\multirow{2}{*}{Dataset} & \multirow{2}{*}{Method} & \multicolumn{3}{c||}{Prediction Performance} & \multicolumn{4}{c}{Fairness}                     \\\cline{3-9}
                         &                         & Accuracy ($\uparrow$)   & F1-score ($\uparrow$)   & AUROC ($\uparrow$)  & $\triangle_{EO}$ ($\downarrow$) & $\triangle_{DP}$ ($\downarrow$) & $\delta_{CF}$ ($\downarrow$) & $R^2$ ($\downarrow$)\\
\hline
Synthetic                   & \mymodel-NS  &   $0.722\pm0.023$         &   $0.726\pm 0.025$         &   $0.794\pm 0.028$   &   $0.061\pm0.044$         &   $0.071\pm0.024$               &   $0.005\pm0.002$            & $0.011\pm 0.006$  \\
            & \mymodel-NN   &     $0.725 \pm 0.026$       &  \bm{$0.727 \pm 0.016$}          &    $0.794 \pm 0.024$  &   $0.066 \pm 0.048$         &  $0.086\pm0.033$                &     $0.008\pm 0.004$          &   $0.016\pm0.005$\\
            & \mymodel-NP & \bm{$ 0.729 \pm 0.022$}            &  \bm{$0.727 \pm 0.027$}          &  \bm{$0.796 \pm 0.016$}    &   $0.094 \pm 0.051$         &      $0.116 \pm 0.063$            &   $0.012 \pm 0.005$            & $0.023 \pm 0.018$  \\
            & \mymodel-NC   &   $0.720 \pm 0.019$         &   $0.725 \pm 0.018$         &  $0.793 \pm 0.018$    &   $0.058 \pm 0.042$         &    $0.069\pm 0.028$              &      $0.006 \pm 0.003$         &  $0.012 \pm 0.004$ \\
            & \mymodel  &  $0.718 \pm 0.018$          &   $0.724 \pm 0.022$         &  $0.793 \pm 0.014$    &    \bm{$0.052 \pm 0.038$}        &  \bm{$0.064 \pm 0.038$}                & \bm{$0.002 \pm 0.002$ }         &  \bm{$0.007 \pm 0.006$}     \\
\hline
Bail                   & \mymodel-NS        & $0.854 \pm 0.020$            &   $0.802 \pm 0.014$         &   $0.897 \pm 0.020$   &   $0.027\pm 0.024$         &    $0.066 \pm 0.020$              &    $0.014\pm 0.007$           & $0.056\pm 0.018$  \\
 & \mymodel-NN    &      $0.855 \pm 0.024$      &  \bm{$0.804 \pm 0.024$}          &  \bm{$0.898 \pm 0.020$}    &   $0.032 \pm 0.027$         &  $0.068\pm 0.023$                &    $0.022 \pm 0.009$           &   $0.058 \pm 0.016$\\
            & \mymodel-NP          &    \bm{$0.860 \pm 0.022$}        &    \bm{$0.804 \pm 0.031$}        &  \bm{$0.898 \pm 0.022$}    &    $0.041 \pm 0.028$        &    $0.073 \pm 0.028$           &       $0.027 \pm 0.010$           &  $ 0.064 \pm 0.019$ \\
            & \mymodel-NC       & $0.853 \pm 0.024$            &   $0.801\pm 0.019$         &  $0.896 \pm 0.021$    &     $0.025\pm 0.027$       &       $0.064\pm 0.014$           &  $0.007 \pm 0.004$             &  $0.053\pm 0.014$ \\
            & \mymodel         &     $0.852 \pm 0.026$       &    $0.800 \pm 0.031$        &  $0.896 \pm 0.016$    &   \bm{$0.019 \pm 0.023$}         &     \bm{$0.058 \pm 0.017$}             & \bm{$0.003 \pm 0.002$}    &   \bm{$0.049 \pm 0.012$}          \\
\hline
Credit                   & \mymodel-NS    & $0.749\pm 0.014$           &    $0.831\pm 0.024$        &   $0.741\pm 0.011$   &   $0.089\pm 0.018$         &      $0.109\pm 0.038$            &     $0.016\pm 0.027$          &  $0.012\pm 0.005$ \\
            & \mymodel-NN    &  $0.751\pm0.012$          &    $0.832\pm 0.018$        &    $0.742\pm 0.009$  &     $0.092\pm 0.034$       &  $0.114 \pm 0.043$                &     $0.020\pm 0.047$          & $0.013 \pm 0.004$   \\
             & \mymodel-NP  &    $0.753 \pm 0.018$        &  \bm{$0.836 \pm 0.017$}          &  \bm{ $0.749 \pm 0.010$}   &   $0.099 \pm 0.043$         & $0.122 \pm 0.049$                 &     $0.028 \pm 0.054$          & $0.016 \pm 0.007$  \\
            & \mymodel-NC    &  $0.749\pm0.015$   &  $0.830\pm0.011$          & $0.741\pm0.009$     &     $0.088\pm 0.012$       &     $0.106\pm0.011$             &  $0.004\pm 0.002$             & $0.013\pm 0.004$  \\
            & \mymodel           &   \bm{$0.755 \pm 0.011$}         &   {$0.835 \pm 0.008$}         &   {$0.740 \pm 0.008$}   &   \bm{$0.086 \pm 0.018$}         & \bm{$0.104 \pm 0.013$}       &   \bm{$0.001 \pm 0.001$}       &  \bm{$0.010 \pm 0.003$}\\
\hline
\end{tabular}
\vspace{-3mm}
\end{table*}

\vspace{-2mm}
\subsection{Experiment Settings}
\noindent\textbf{Metrics.} We evaluate the proposed framework with respect to two aspects: prediction performance and fairness. To evaluate the prediction performance, we use the widely-used node classification metrics: accuracy, F1-score, and AUROC. To measure the fairness of the representations, we first use two metrics which are commonly used in statistical fairness:
$
    \triangle_{SP} = |P(\hat{Y}_i|S_i=0)-P(\hat{Y}_i|S_i=1)|, 
$
and 
$
    \triangle_{EO} = |P(\hat{Y}_i|Y_i=1,S_i=0)-P(\hat{Y}_i|Y_i=1,S_i=1)|. 
$
To evaluate graph counterfactual fairness, we design a metric $\delta_{CF}$:
\begin{equation}
     \delta_{CF} = |P((\hat{Y}_i)_{{S}\leftarrow \mathbf{s}'}|X=\mathbf{X},A=\mathbf{A})-P((\hat{Y}_i)_{{S}\leftarrow \mathbf{s}''}|X=\mathbf{X},A=\mathbf{A})|,
\end{equation}
where $\mathbf{s}',\mathbf{s}''\in \{0,1\}^n$ are arbitrary values of sensitive attribute of all nodes.
As there are too many different counterfactuals (e.g., there are $2^n$ cases for a graph with $n$ nodes), it is difficult to evaluate the difference of predictions under all these counterfactuals. Therefore, we evaluate the graph counterfactual fairness of the proposed model in the following way: on each dataset, we control the rate of sensitive subgroup population and randomly perturb the sensitive attribute of all nodes. More specifically, we randomly select $0\%, 50\%, 100\%$ nodes, and set their sensitive attribute values to be $1$, while set the sensitive attribute of other nodes to be $0$. With such perturbations, we generate counterfactual data for the whole graph with different ratios of sensitive subgroup, based on the causal model described in Section \ref{sec:dataset}. Intuitively, these perturbations implicitly control the distribution of the sensitive attribute in each node's neighborhood, and we take the averaged ratio of nodes which flip their predicted labels as an estimation for $\delta_{CF}$. Besides, we also compute the R-square $R^2(\hat{Y}_i, \tilde{{S}}_i)$ to measure how well a linear regression predictor for $\hat{Y}_i$ can be explained by the summary of the neighboring sensitive attribute values for any node $i$. Here we use the mean aggregator over the sensitive attribute values of all one-hop neighbors and each node $i$ itself to compute the sensitive attribute summary $\tilde{{S}}_i$. This R-square metric can reflect the statistical dependency between $\hat{Y}_i$ and $\tilde{{S}}_i$.

\noindent\textbf{Baselines.} We compare the proposed framework with several  state-of-the-art node representation learning methods. We divide them into two categories: 1) node representation learning methods without fairness constraints: these methods only aim to encode useful information from the input graph and improve the prediction performance in downstream tasks. We use graph convoluntional network (\textbf{GCN}) \cite{kipf2016semi}, \textbf{GraphSAGE} \cite{hamilton2017inductive}, and Graph Isomorphism Network (\textbf{GIN}) \cite{xu2018powerful} as baselines; 2) fair representation learning methods on graphs: these methods target on learning fair node representation on graphs. Among them, \textbf{C-ENC} \cite{bose2019compositional} and \textbf{FairGNN} \cite{dai2020fairgnn} enforces fairness with an adversarial discriminator to predict the sensitive attribute; \textbf{NIFTY} \cite{agarwal2021towards} enforces fairness by maximizing the similarity of representations learned from the original graph and their augmented counterfactual graphs, where the sensitive attribute values of all nodes are flipped, while other parts remain unchanged. We use two variants of it with GCN or GraphSAGE encoders, denoted by \textbf{NIFTY-GCN} and \textbf{NIFTY-SAGE}, respectively.

\noindent\textbf{Setup.} Each dataset is randomly splitted into $60\%/20\%/20\%$ training/validation/test set. Unless otherwise specified, we set the hyperparameters as $\lambda=0.6$, $C=2$, $\lambda_s=0.4$, $\beta=10$, $\mu=1e-5$, $k=20$, $B=4$. The learning rate is $0.001$, the number of epochs is $1,000$, the representation dimension is $1,024$, batch size is $100$. Experimental results are averaged over ten repeated executions. We use the Adam optimizer and implement our method with Pytorch \cite{paszke2019pytorch}.

\begin{figure}[t]
\centering
  \begin{subfigure}[b]{0.232\textwidth}
        \centering
        \includegraphics[height=1.in]{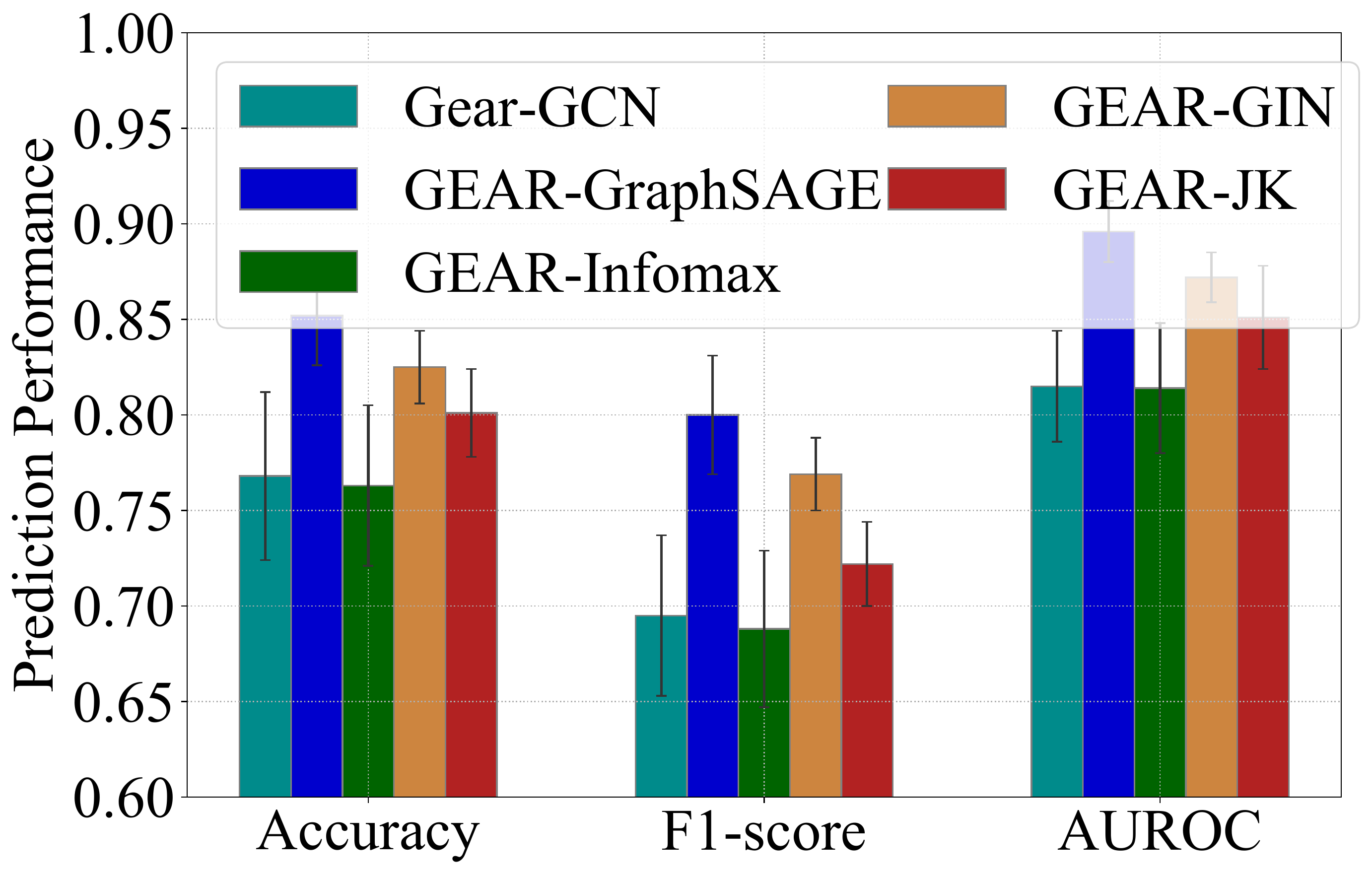}
        \vspace{-2mm}
        \caption{Prediction}
    \end{subfigure}
  \begin{subfigure}[b]{0.232\textwidth}
        \centering
        \includegraphics[height=1.in]{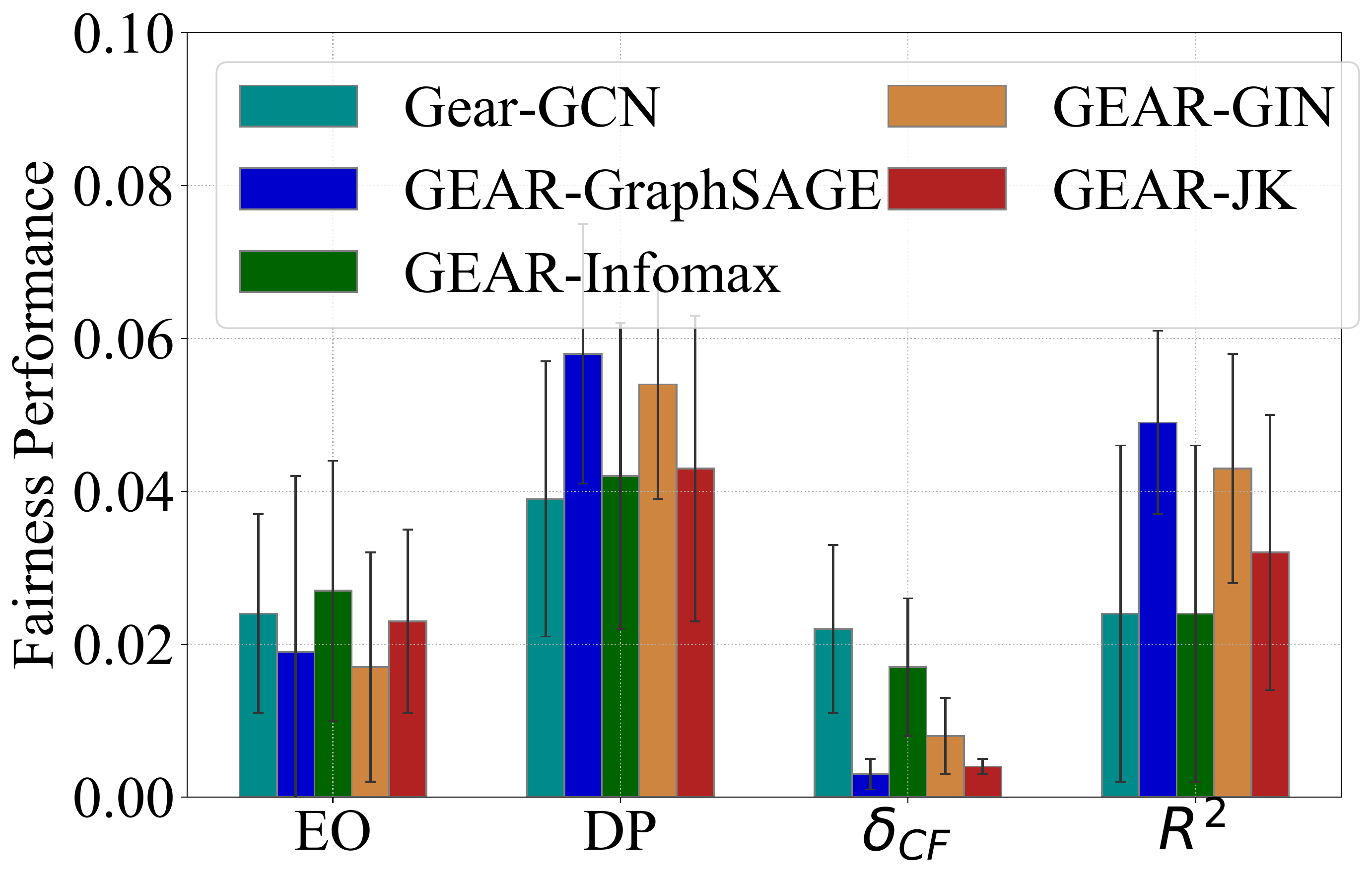}
        \vspace{-2mm}
        \caption{Fairness}
    \end{subfigure}
    \vspace{-2mm}
  \caption{Comparison of the performance of different subgraph encoders in GEAR on Bail dataset.}
  \label{fig:encoder}
  \vspace{-3mm}
\end{figure}

\begin{figure}[t]
\centering
  \begin{subfigure}[b]{0.22\textwidth}
        \centering
        \includegraphics[height=1.in]{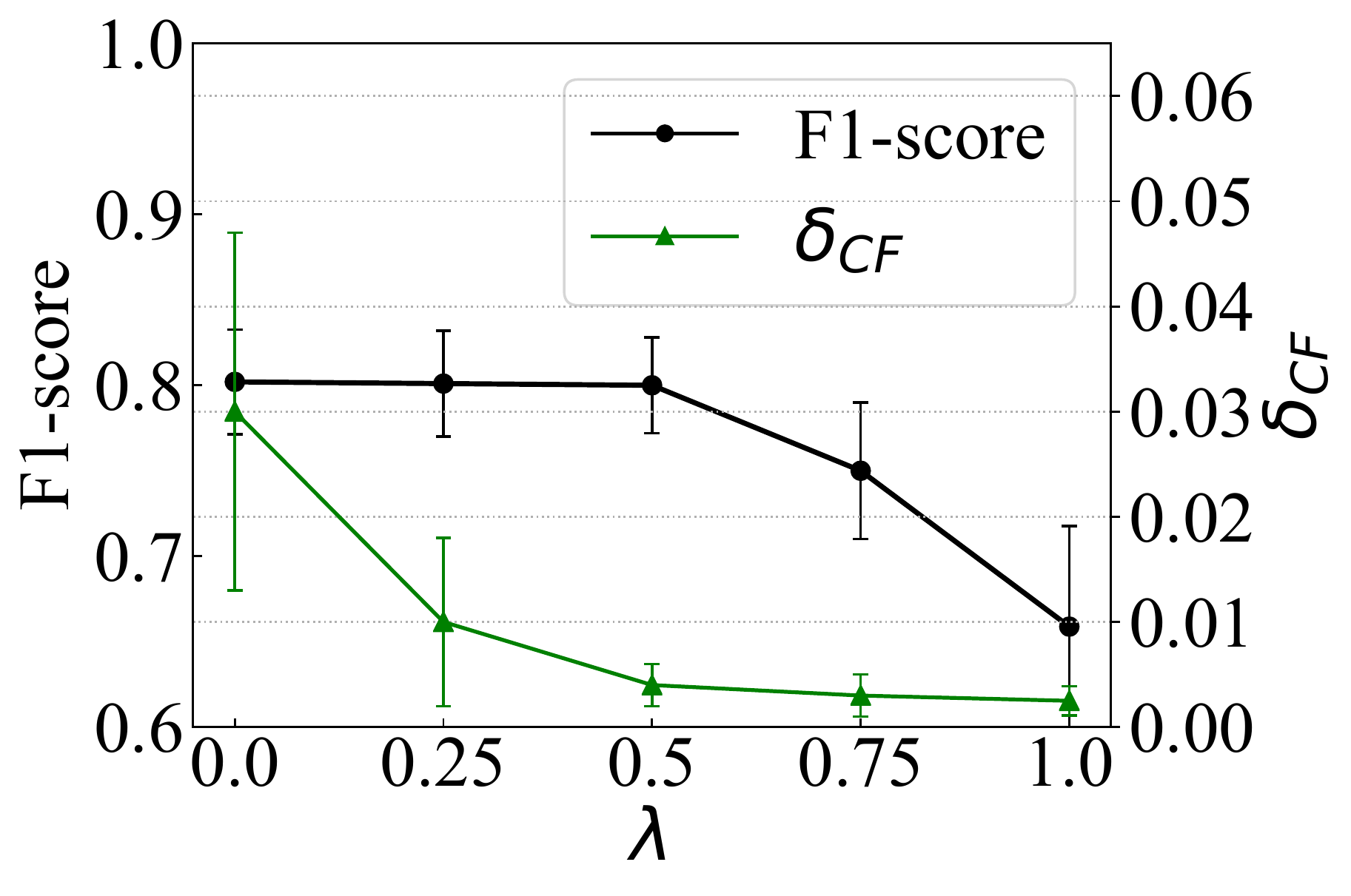}
        \vspace{-2mm}
        \caption{$\lambda$}
    \end{subfigure}
  \begin{subfigure}[b]{0.232\textwidth}
        \centering
        \includegraphics[height=1.in]{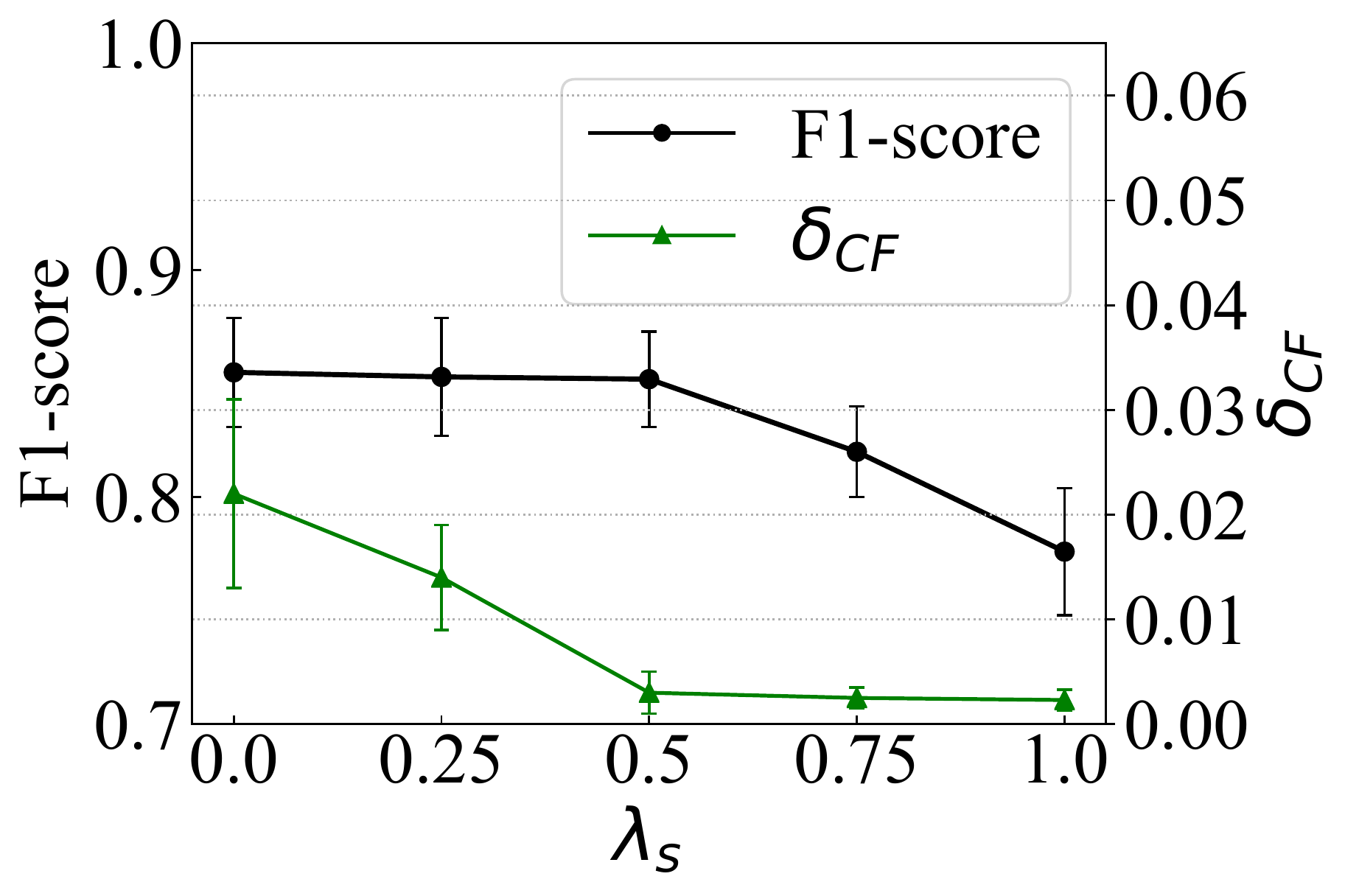}
        \vspace{-2mm}
        \caption{$\lambda_s$}
    \end{subfigure}
    \begin{subfigure}[b]{0.22\textwidth}
        \centering
        \includegraphics[height=1.in]{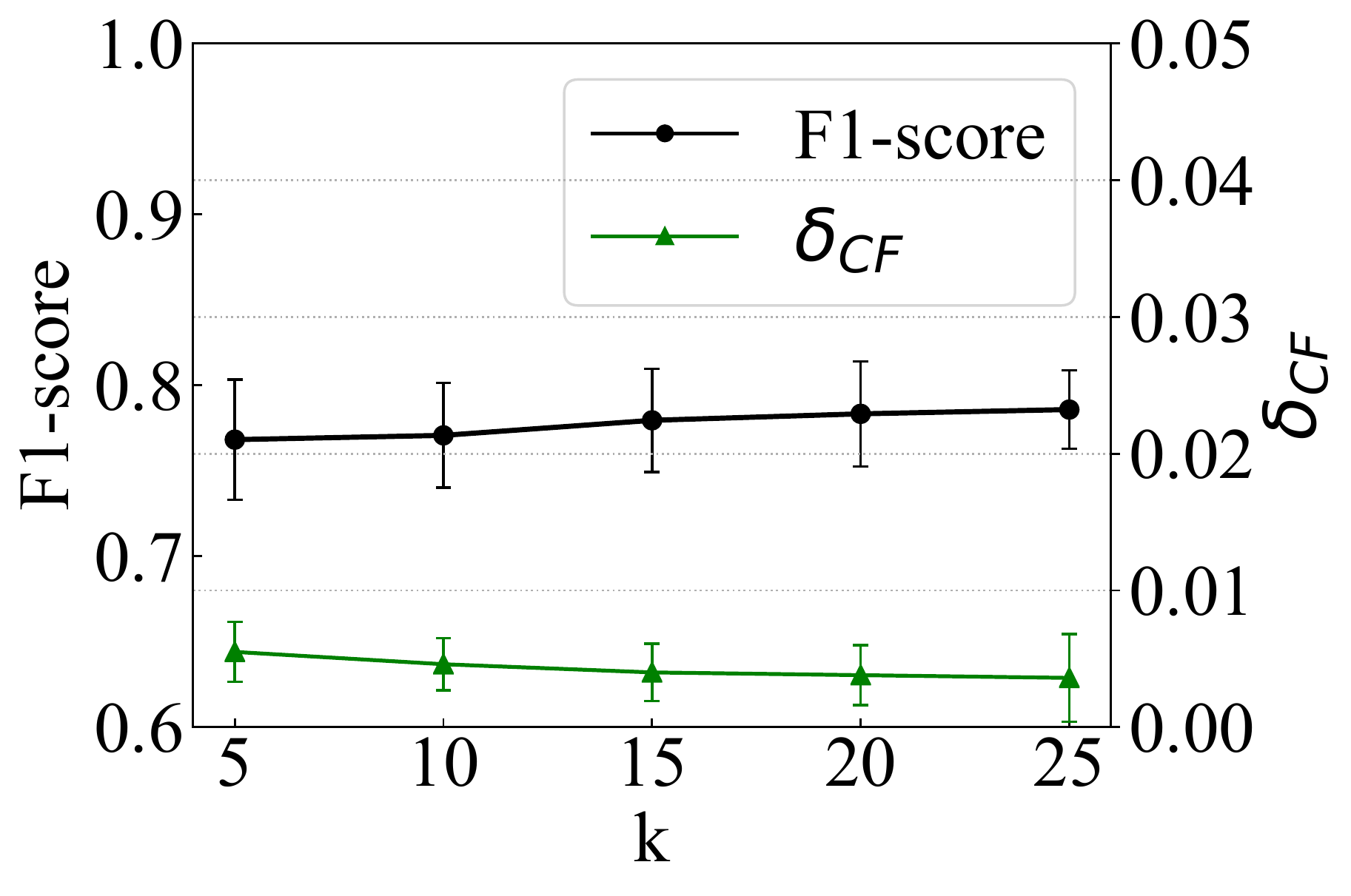}
        \vspace{-2mm}
        \caption{Subgraph size $k$}
    \end{subfigure}
  \begin{subfigure}[b]{0.232\textwidth}
        \centering
        \includegraphics[height=1.in]{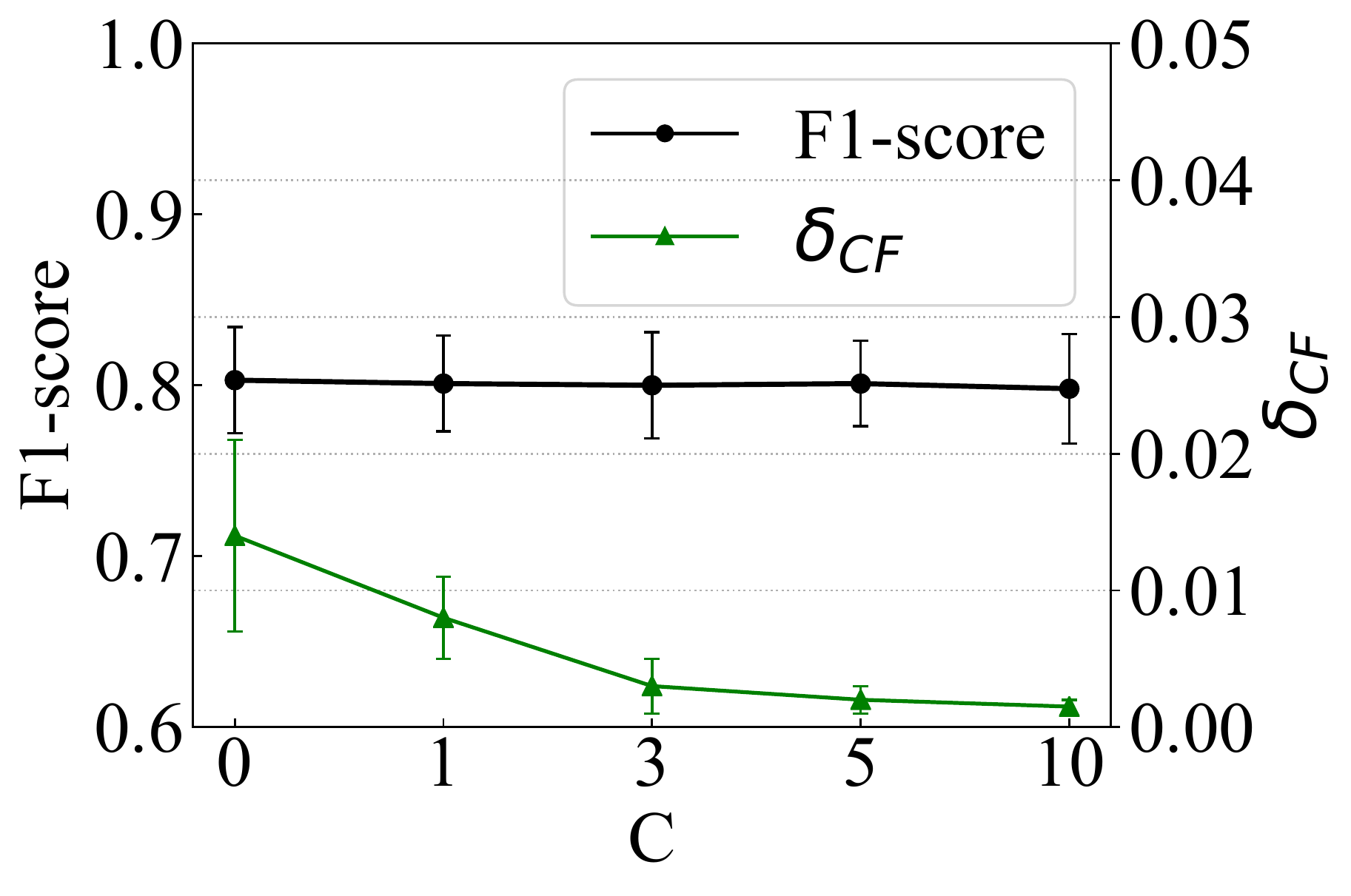}
        \vspace{-2mm}
        \caption{\# of neighbor-perturbations $C$}
    \end{subfigure}
        \vspace{-2mm}
  \caption{Parameter study on Bail dataset.}
  \label{fig:parameter}
  \vspace{-3mm}
\end{figure}

\vspace{-2mm}
\subsection{Prediction Performance and Fairness}
The performance of prediction and fairness is shown in Table~\ref{tab:baseline}. The best results are shown in \textbf{bold}, and the runner-up results are \underline{underlined}. Generally speaking, we have the following observations: 
1) The proposed model \mymodel~ shows comparable prediction performance with the state-of-the-art node representation learning methods, and it outperforms all the fair node representation learning methods in prediction; 
2) The proposed model outperforms all the other fair node representation learning methods in $\delta_{CF}$ and $R^2$. 
These two fairness metrics explicitly consider the causal/statistical relation between the neighboring sensitive attribute and the model prediction, thus this observation validates the effectiveness of our framework in mitigating the biases from neighbors. 
Besides, \mymodel~ also performs well in other fairness metrics $\triangle_{EO}$ and $\triangle_{DP}$.
The baseline NIFTY also has good performance in graph counterfactual fairness, because NIFTY also generates counterfactuals during training. Although NIFTY does not explicitly consider the causal effect from neighbors' sensitive attribute on each node, its counterfactuals still implicitly promote graph counterfactual fairness. However, our method still outperforms all these fairness methods mainly for two reasons: a) \mymodel~ generates multiple versions of counterfactuals with self-perturbation and neighbor-perturbation. It has better coverage of the space of possible counterfactuals, while NIFTY only generates one counterfactual by flipping all nodes' sensitive attribute values, here the influence from the neighbors' sensitive attribute may counteract with each other; b) \mymodel~ generates counterfactuals which include changes in both features and graph structure after modifying the sensitive attribute, rather than simply changing the sensitive attribute. More specifically, the counterfactual augmentation component in \mymodel~ removes biases caused by misusing the descendants of the sensitive attribute in node representation learning. 

\vspace{-2mm}
\subsection{Model Structure \& Parameter Study}
To investigate the model performance under different options of model structure, we vary the encoder of subgraphs, including GCN \cite{kipf2016semi}, GraphSAGE (mean aggregator) \cite{hamilton2017inductive}, Informax \cite{velickovic2019deep}, GIN \cite{xu2018powerful}, and Jumping Knowledge (JK) \cite{xu2018representation}. 
Due to the space limit, we only show the results in Bail in Fig.~\ref{fig:encoder}, but the observations are consistent in other datasets. Generally, all the subgraph encoders show good performance in both prediction and fairness.
Among them, GraphSAGE encoder shows the best performance over all the variants.

To evaluate the robustness of \mymodel~ under different parameters, we vary the parameters $\lambda$ (the weight of constraint for graph counterfactual fairness), $\lambda_s$ (the weight for neighbor-perturbation), subgraph size $k$, and the number of neighbor-perturbations $C$ in $\mathtt{SMP}(\cdot)$ to investigate how the model performance varies. The results over different settings of parameters in Bail are shown in Fig~\ref{fig:parameter}. We observe that: 1) The model achieves better fairness with larger $\lambda$ and $\lambda_s$ while with slight sacrifice of the prediction performance; 2) The model achieves better fairness with relatively larger subgraph size, but this improvement becomes less significant when the subgraph size is over $20$. 3) When the number of neighbor-perturbations in $\mathtt{SMP}(\cdot)$ becomes larger, the predictions become more fair. These observations generally match our expectation. 

\vspace{-2mm}
\subsection{Ablation Study}

In ablation study, we compare different variants of \mymodel~ to verify the effectiveness of different components. We first remove self-perturbations, denote this variant as \mymodel-NS. Next, we remove neighbor-perturbations, denoted by \mymodel-NN. We then remove all the perturbations, denote this variant as \mymodel-NP. We remove the counterfactual data augmentation module, just flip the sensitive attribute values, and denote this variant as \mymodel-NC. The model performance of these variants is shown in Table~\ref{tab:ablation}. We observe that all the variants perform worse than \mymodel~ with respect to fairness. These results validate the effectiveness of different components in \mymodel~ for learning fair node representations. 

\vspace{-2mm}
\section{Related Work}

\noindent\textbf{Graph representation learning.}
Many efforts have been made for graph representation learning \cite{hamilton2020graph,wang2018graphgan,fey2019fast} in recent years. Among them, neural network methods encode the attributes and graph structure 
into a latent space as representations to capture useful information. 
These methods include the well-known graph convolutional networks (GCNs) \cite{kipf2016semi}, variational graph auto-encoders (VGAE) \cite{kipf2016variational}, GraphSAGE \cite{hamilton2017inductive}, and structural deep network embedding (SDNE) \cite{wang2016structural}. 
Recently, subgraph-based methods \cite{jiao2020sub,nguyen2018learning,crouse2019improving} utilize the  correlation between central nodes and their sampled subgraphs to capture regional structure information and improve the model scalability. Despite the success of these methods in different domains, they may exhibit biases against certain sensitive groups.

\noindent\textbf{Fairness on graphs.}
Fairness in machine learning has attracted significant attention recently \cite{verma2018fairness,kwiatkowska1989survey,binns2020apparent}. Typical fairness notions include group fairness  \cite{zemel2013learning,hardt2016equality,zafar2017fairness,dieterich2016compas,chouldechova2017fair}, individual fairness \cite{dwork2012fairness,sharifi2019average}, and counterfactual fairness \cite{kusner2017counterfactual,russell2017worlds,chiappa2019path,wu2019counterfactual}.
Recent works \cite{bose2019compositional,buyl2020debayes,dai2020fairgnn,agarwal2021towards,kang2020inform,dong2021individual} 
promote fairness 
in node representation learning. Most of works are based on adversarial learning \cite{beutel2017data,zhang2018mitigating}, aiming to prevent the learned representations from accurately predicting the corresponding sensitive attribute. These works focus on removing the statistical dependency between the sensitive attribute and predictions elicited by the learned representations, but do not consider the biases in the features, graph structure or labels due to the causal effect from the sensitive attribute on them. Differently, few works \cite{agarwal2021towards,agarwal2021towards2} extend counterfactual fairness to graphs. 
However, most of these works do not consider the potential biases brought by the sensitive attribute of neighboring nodes, and the 
causal effect from the sensitive attribute to other node features and graph structure.



\section{Conclusion}
In this paper, we propose a novel fairness notion of graph counterfactual fairness, which explicitly considers the causal influence from the neighboring nodes' sensitive attribute to each node, as well as the causal effect from the sensitive attribute to other features and the graph structure. We propose a novel framework \mymodel~ to learn node representations which can achieve graph counterfactual fairness and good prediction performance simultaneously. Specifically, in \mymodel, we use a counterfactual data augmentation module to generate counterfactuals with interventions on the sensitive attribute of different nodes. 
\mymodel~ then maximizes the similarity between the node representations learned from the original data and different counterfactuals. Experimental results on synthetic and real-world graphs validate the effectiveness of our framework with respect to both prediction performance and fairness. 

\vspace{-1mm}
\section*{Acknowledgements}
This material is supported by the National Science Foundation (NSF) under grants \#1955151, \#1934600, \#2006844, a JP Morgan Chase Faculty Research Award, and a Cisco Faculty Research Award. 

\clearpage
\bibliographystyle{ACM-Reference-Format}
\balance
\bibliography{ref}

\end{document}


\title{Supplementary Material}

\author{Jing Ma}
\affiliation{%
  \institution{University of Virginia}
}
\email{jm3mr@virginia.edu}
\renewcommand{\shortauthors}{}

\newcommand{\mymodel}{GEAR}
\newcommand{\bigCI}{\mathrel{\text{\scalebox{1.07}{$\perp\mkern-10mu\perp$}}}}

\maketitle


\section{Implementation Details}
In our model, we set the order of neighbor nodes as $10$, dropout rate as $0.5$, the subgraph encoders are implemented based on the torch geometric package. The prediction $f(\cdot)$ over the learned representations is implemented with a two-layer multilayer perceptron (MLP) with BatchNorm, and we use ReLU as activation function.

\section{More Experimental Results}
In Table \ref{tab:encoder}, we compare the performance of different subgraph encoders in the proposed framework \mymodel~ on all the datasets. Generally, the observations are consistent as the results on the bail dataset as shown in Fig. 4.

\begin{table*}[t]
\small
\centering
 \caption{Comparison of the performance of different subgraph encoders in \mymodel.}
 \vspace{-2mm}
 \label{tab:encoder}
 \begin{tabular}{l||l||c|c|c||c|c|c|c}
 \hline
\multirow{2}{*}{Dataset} & \multirow{2}{*}{Method} & \multicolumn{3}{c||}{Prediction} & \multicolumn{4}{c}{Fairness}                     \\\cline{3-9}
                         &                         & Accuracy ($\uparrow$)   & F1-score ($\uparrow$)   & AUROC ($\uparrow$)  & $\triangle_{EO}$ ($\downarrow$) & $\triangle_{DP}$ ($\downarrow$) & $\delta_{CF}$ ($\downarrow$) & $R^2$ ($\downarrow$)\\
\hline
Synthetic                   & \mymodel-GCN   &     $0.675 \pm 0.018$       &  $0.680 \pm 0.031$          &   $0.749 \pm 0.017$   &  $0.045 \pm 0.036$          &    $0.051 \pm 0.040$              &   $0.003 \pm 0.002$            & $0.010 \pm 0.012$  \\
            & \mymodel-GraphSAGE            &  $0.718 \pm 0.018$          &   $0.724 \pm 0.022$         &  $0.793 \pm 0.014$    &    $0.052 \pm 0.038$        &  $0.064 \pm 0.038$                & \bm{$0.002 \pm 0.002$ }         &  $0.007 \pm 0.006$     \\
            & \mymodel-Infomax     & $0.669 \pm 0.019$   &    $0.671 \pm 0.035$        & $0.746 \pm 0.018$     &   $0.053 \pm 0.044$         &  $0.056 \pm 0.048$                &   $0.003 \pm 0.001$            &  $0.010 \pm 0.016$ \\
            & \mymodel-GIN &   $0.676 \pm 0.015$         &   $0.677 \pm 0.026$         &  $0.741 \pm 0.020$    &   $0.033 \pm 0.025$         & $0.055 \pm 0.027$                 &  $0.015 \pm 0.009$             &  $0.009 \pm 0.009$ \\
            & \mymodel-JK   & $0.682 \pm 0.016$            &  $0.693 \pm 0.025$          &  $0.747 \pm 0.018$    &    $0.065 \pm 0.040$        &               $0.068 \pm 0.050$   &    \bm{$0.002 \pm 0.001$}           &   $0.013 \pm 0.013$\\
\hline
Bail                   & \mymodel-GCN      &   $0.768 \pm 0.044$         &       $0.695 \pm 0.042$     &  $0.815 \pm 0.029$    &  $0.024 \pm 0.013$          &    $0.039 \pm 0.018$              &  $0.022 \pm 0.011$       & $0.024 \pm 0.022$   \\
            & \mymodel-GraphSAGE                    &     $0.852 \pm 0.026$       &    $0.800 \pm 0.031$        &  $0.896 \pm 0.016$    &   $0.019 \pm 0.023$         &     $0.058 \pm 0.017$             & \bm{$0.003 \pm 0.002$}    &   $0.049 \pm 0.012$          \\
            & \mymodel-Infomax  &  $0.763 \pm 0.042$          &  $0.688 \pm 0.041$          &  $0.814 \pm 0.034$    &  $0.027 \pm 0.017$          &    $0.042 \pm 0.020$              &   $0.017 \pm 0.009$            & $0.024 \pm 0.022$  \\
            & \mymodel-GIN    &  $0.825 \pm 0.019$           &  $0.769 \pm 0.019$          &   $0.872 \pm 0.013$   &    $0.017 \pm 0.015$        &      $ 0.054 \pm 0.015$            &   $0.008 \pm 0.005$            &  $0.043 \pm 0.015$ \\
            & \mymodel-JK  &    $0.801 \pm 0.023$        &  $0.722 \pm 0.022$          &   $0.851 \pm 0.027$   &   $0.023 \pm 0.012$         & $0.043 \pm 0.020$                 &    $0.004 \pm 0.001$           & $0.032 \pm 0.018$  \\
\hline
Credit                   & \mymodel-GCN   &  $0.718 \pm 0.013$          &   $0.811 \pm 0.012$         &   $0.684 \pm 0.006$   &    $0.074 \pm 0.023$        &      $0.088 \pm 0.021$            &     \bm{$0.001 \pm 0.001$}          & $0.014 \pm 0.003$  \\
            & \mymodel-GraphSAGE           &   \bm{$0.755 \pm 0.011$}         &   \bm{$0.835 \pm 0.008$}         &   \bm{$0.740 \pm 0.008$}   &   $0.086 \pm 0.018$         & {$0.104 \pm 0.013$}       &   \bm{$0.001 \pm 0.001$}       &  \bm{$0.010 \pm 0.003$}\\
            & \mymodel-Infomax   & $0.723 \pm 0.011$           &   $0.816 \pm 0.010$         &  $0.685 \pm 0.006$    &    $0.074 \pm 0.020$        &       $0.087 \pm 0.018$           &     \bm{$0.001 \pm 0.001$}          &  $0.013 \pm 0.003$ \\
            & \mymodel-GIN &  $0.720 \pm 0.017$          &  $0.810 \pm 0.015$          &  $0.707 \pm 0.006$    &   $0.067 \pm 0.027$         &    $0.081 \pm 0.028$              &    $0.005 \pm 0.002$           &  $0.012 \pm 0.006$  \\
            & \mymodel-JK   & $0.745 \pm 0.016$           &  $0.812 \pm 0.026$          &  $0.688 \pm 0.015$    &   $0.057 \pm 0.022$         &  $0.079 \pm 0.020$                &   $0.002 \pm 0.001$            & $0.008 \pm 0.002$  \\
\hline
\end{tabular}
\vspace{-3mm}
\end{table*}